\documentclass[10pt,twocolumn,letterpaper]{article}

\usepackage[pagenumbers]{iccv} 

\usepackage{newfloat}
\usepackage{listings}
\usepackage{multirow}
\usepackage{tcolorbox}
\usepackage{algorithm}
\usepackage{algorithmic}
\usepackage{colortbl}
\usepackage{lscape}
\usepackage{graphicx}
\usepackage{booktabs}
\usepackage{pifont}

\definecolor{iccvblue}{rgb}{0.21,0.49,0.74}
\usepackage[pagebackref,breaklinks,colorlinks,allcolors=iccvblue]{hyperref}
\usepackage{fontawesome5} 
\usepackage{hyperref}
\usepackage{xcolor}

\def\paperID{1276}
\def\confName{ICCV}
\def\confYear{2025}
\usepackage{marvosym} 

\definecolor{c1}{HTML}{F2C335}
\definecolor{c2}{HTML}{D9666F}

\title{\textcolor{c1}{Anime}\textcolor{c2}{Gamer}: Infinite Anime Life Simulation with Next Game State Prediction}

\author{Junhao Cheng$^{1,2}$ \and Yuying Ge$^{1,\textsuperscript{\Letter}}$ \and Yixiao Ge$^{1}$ \and Jing Liao$^{2}$ \and Ying Shan$^{1}$ \and \vspace{-3mm} \\ \centering $^{1}$ARC Lab, Tencent PCG \ \ \ \ $^{2}$City University of Hong Kong \and \vspace{-5mm} \\ \centering \url{https://howe125.github.io/AnimeGamer.github.io/}}

\begin{document}

\maketitle

\begin{abstract}
Recent advancements in image and video synthesis have opened up new promise in generative games. One particularly intriguing application is transforming characters from anime films into interactive, playable entities. This allows players to immerse themselves in the dynamic anime world as their favorite characters for life simulation through language instructions. Such games are defined as ``infinite game'' since they eliminate predetermined boundaries and fixed gameplay rules, where players can interact with the game world through open-ended language and experience ever-evolving storylines and environments. Recently, a pioneering approach for infinite anime life simulation employs large language models (LLMs) to translate multi-turn text dialogues into language instructions for image generation. However, it neglects historical visual context, leading to inconsistent gameplay. Furthermore, it only generates static images, failing to incorporate the dynamics necessary for an engaging gaming experience. In this work, we propose AnimeGamer, which is built upon Multimodal Large Language Models (MLLMs) to generate each game state, including dynamic animation shots that depict character movements and updates to character states, as illustrated in Figure~\ref{fig: head}. We introduce novel action-aware multimodal representations to represent animation shots, which can be decoded into high-quality video clips using a video diffusion model. By taking historical animation shot representations as context and predicting subsequent representations, AnimeGamer can generate games with contextual consistency and satisfactory dynamics. Extensive evaluations using both automated metrics and human evaluations demonstrate that AnimeGamer outperforms existing methods in various aspects of the gaming experience. Codes and checkpoints are available at~\url{https://github.com/TencentARC/AnimeGamer}.
\end{abstract}

\section{Introduction}

Recent advances in generative models have significantly enhanced anime production, particularly in character design and the creation of character-centric images and videos~\cite{guo2023animatediff,yang2024seed,xu2024dreamanime}. This progress raises an intriguing question: Can we transcend static content generation to \textit{create infinite anime games by transforming characters from anime films into interactive, playable entities?} Imagine experiencing the life of characters crafted by Hayao Miyazaki within a dynamic anime world. Users can continuously interact with this world using open-ended language instructions, while the model consistently generates game states. These states encompass dynamic animation shots and updates to character attributes such as stamina, social, and entertainment values, as illustrated in Figure~\ref{fig: head}.

The concept of ``Anime Life Simulation" falls under the category of infinite games, as explored in recent research~\cite{li2024unbounded}. In these games, all behaviors and graphics are generated through AI models, eliminating the need for predefined game rules and pre-designed graphics. By leveraging the capabilities of generative models, we can create immersive and ever-evolving gaming experiences that allow players to experience the lives of their favorite anime characters in unprecedented ways.

\begin{figure}[!t]
  \centering
  \setlength{\tabcolsep}{1pt} 
  \begin{tabular}{c}
    \includegraphics[width=.49\textwidth]{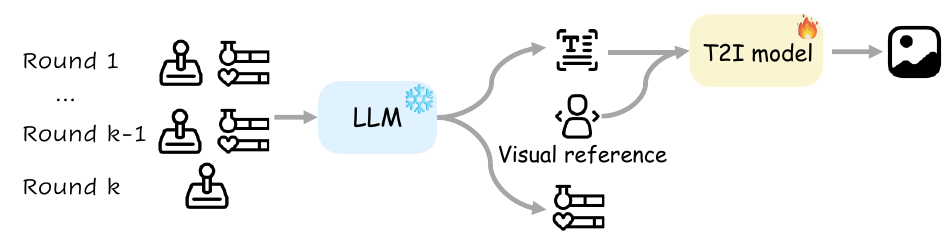} \\
    \multicolumn{1}{c}{\vspace{-12pt}} \\
    \small (a) LLM-based methods (e.g., Unbounded~\cite{li2024unbounded}) \\
    \multicolumn{1}{c}{\vspace{-8pt}} \\
    \includegraphics[width=.49\textwidth]{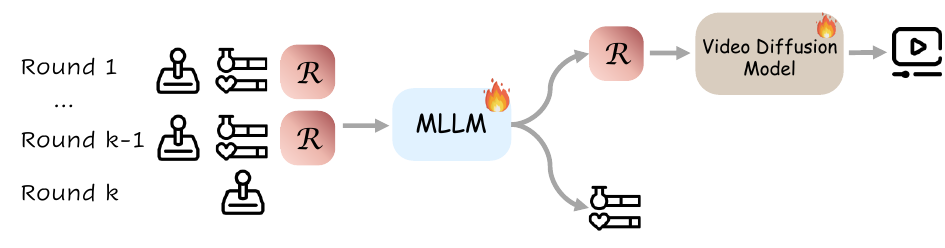} \\
    \multicolumn{1}{c}{\vspace{-12pt}} \\
    \small \textbf{(b) AnimeGamer} \\
  \end{tabular}
  \caption{Comparison of AnimeGamer with previous methods. Unbounded employs an LLM to translate multi-turn \textbf{text-only} dialogues into language descriptions for \textbf{static image} generation, with an additional condition based on reference images. In contrast, AnimeGamer utilizes an MLLM to predict multimodal representations $\mathcal{R}$ by incorporating historical \textbf{multimodal} context as input. These generated representations can then be directly decoded into consistent \textbf{dynamic clips} using a video diffusion model.}
   \label{fig: intro comparison}
\end{figure}

Some recent works utilize generative models to generate the next frame in existing games~\cite{oasis,yang2024playable,valevski2024diffusion} or open-domain game scenarios~\cite{bruce2024genie,che2024gamegen,yu2025gamefactory} by taking the previous game frames and user controls (mouse or keyboard) as the input. However, these approaches are constrained by limited command inputs (\textit{e.g.}, directional controls) and exploration within predefined environments, which categorizes them as finite games. The pioneering work Unbounded~\cite{li2024unbounded} addresses the challenge of infinite anime life simulation by employing an LLM as a router to translate multi-turn text-only dialogues into language captions for static image generation, as illustrated in Figure~\ref{fig: intro comparison}. However, this approach neglects historical visual context, which is crucial for maintaining continuity and coherence in gameplay. Furthermore, it is limited to generating static images, which fails to represent the dynamic interactions and movements essential for an engaging gaming experience (imagine a game world where characters remain completely motionless).

To overcome the above limitations, in this work, we propose AnimeGamer, which leverages an MLLM to generate game states for infinite anime life simulation. We introduce novel action-aware multimodal representations, which effectively capture the intricacies of animation shots. These representations can be seamlessly decoded into high-quality video clips using a video diffusion model. By utilizing historical multimodal representations and character state updates as input, AnimeGamer can predict subsequent game states, ensuring that the generated animation shots are contextually consistent with satisfactory dynamics, and update character states reasonably. In addition, we propose an automatic data collection pipeline from anime films, empowering players to experience the life of their favorite characters in an infinite game through training models on their customized data. To evaluate the effectiveness of our model, we tailor related SOTA methods to this task and design evaluation metrics including both automated and human assessments. The evaluation results demonstrate that our model performs favorably in terms of instruction following, contextual consistency and overall gaming experience.

We make the following contributions in this work: 
\begin{itemize}
    \item We propose AnimeGamer for infinite anime life simulation. Powered by an MLLM, our model takes multimodal context as input to predict the next game state, including dynamic animation shots and character state updates, providing an immersive gaming experience.
    \item We introduce novel action-aware multimodal representations to represent animation shots, which can be decoded into high-quality video clips using a video diffusion model. By taking multimodal representations as input to predict subsequent representations, our approach ensures contextual consistency and satisfactory dynamics throughout gameplay.
    \item We conduct both qualitative and quantitative evaluations, including user studies, to demonstrate the effectiveness of AnimeGamer.
\end{itemize}

\section{Related Works}
\vspace{1mm} \noindent {\bf Generative Games}

\vspace{1mm} \noindent \textit{Finite Games Generation.} Finite games are defined by James P. Carse as games that are played for the purpose of winning with boundaries, fixed rules, and a definitive endpoint~\cite{carse1986finite}. Existing efforts for generative finite games can be primarily categorized into two main approaches: partly generated and fully generated. The partly generated methods rely on pre-existing games or a hard-coded systems, with AI assisting in generating specific game components. Some efforts have focused on using AI to design game interfaces~\cite{gaudl2018exploring,eladhari2011ai} or develop AI-driven games~\cite{treanor2015ai,ennabili2023comparison}. Other approaches have been explored to generate dynamic game content~\cite{shaker2016procedural,summerville2018procedural,smith2014future,liu2021deep} or rules~\cite{khalifa2017general} with concept maps~\cite{treanor2012game}, conceptual expansion~\cite{guzdial2021conceptual,guzdial2018automated}, Markov Chains~\cite{snodgrass2014experiments,snodgrass2016learning}, Bayes Nets~\cite{guzdial2016game}, and LSTMs~\cite{summerville2016super,sarkar2018blending}. Some works attempt to leverage the advantage of generative models to create game levels environments based on GANs~\cite{volz2018evolving,kumaran2019generating,schubert2021toad,kim2020learning} or diffusion models~\cite{sun2023language,zhou2024eyes}. While some recent researches try to use LLM or MLLM to design and generate the game mechanics and environments~\cite{sudhakaran2024mariogpt,todd2023level,nasir2023practical,hu2024game,zala2024envgen,anjum2024ink,chung2024patchview}, or act as an agent to take part in playing and simulation~\cite{zheng2023steve,kaiya2023lyfe,park2023generative,fan2022minedojo,liu2024rl}. However, these methods are limited by their reliance on pre-defined, hard-coded systems and rules, which may potentially stifle innovation. On the other hand, the fully generated approaches use AI to create all aspects of game behavior. These efforts mainly focus on replicating specific scenes from existing games such as Minecraft~\cite{oasis}, Mario~\cite{yang2024playable}, and DOOM~\cite{valevski2024diffusion}, or on open-domain scenarios~\cite{bruce2024genie,che2024gamegen,yu2025gamefactory}. However, they only support limited commands within a predefined environment, thus cannot generalize to perform infinite games.

\vspace{1mm} \noindent \textit{Infinite Games Generation.} Carse defines infinite games as those ``played for the purpose of continuing the play"~\cite{carse1986finite}. The latest work Unbounded~\cite{li2024unbounded} introduces the concept of generative infinite game, with an LLM to generate text responses and a pre-trained T2I model enhanced with LoRA~\cite{hu2021lora} for character-consistent image generation. However, since LLM takes only in-context text information as input, this can lead to a decrease in visual coherence in the final generated results. This issue becomes more pronounced when the images need to be further converted into video outputs. In our work, we utilize an MLLM to predict multimodal representations by incorporating historical multimodal context as input.

\vspace{1mm} \noindent {\bf Multi-turn Image\&Video Generation.} Multi-turn T2I and T2V generation require models to generate coherent visual outputs based on human instructions for various applications such as content design and storytelling~\cite{luo2025objectisolatedattentionconsistent,he2025dreamstoryopendomainstoryvisualization}. Benefiting from the in-context learning and generation capabilities~\cite{brown2020language} of LLM and MLLM, existing approaches are usually driven by them and can be primarily categorized into off-the-shelf approaches and end-to-end methods. The former utilize a pre-trained LLM as a router to transform dialogue into character information~\cite{cheng2024theatergen,long2024videodrafter,li2024unbounded}, layouts~\cite{wang2024autostory,cheng2024autostudio}, or captions~\cite{lai2023mini,cao2024visdiahalbench} for generative models. These approaches ignore in-context visual information when generating next game states. As a result, they underperform in terms of contextual consistency and visual coherence. The latter methods~\cite{yang2024seed,zhao2024moviedreamer,xiang2024pandora} leverage MLLM for end-to-end generation. They take account of both text and visual in-context information to predict next image features. However, when applied to infinite anime life simulation that require video output, these methods necessitate an additional video conversion process, which can disrupt in-context coherence and entail additional computational and time costs. In our work, AnimeGamer predicts the next animation shot representations, which can then be decoded into videos with controllable character movements and motion scope.

\section{Methods}

\begin{figure*}[!t]
	\centering
	\includegraphics[width=1\textwidth]{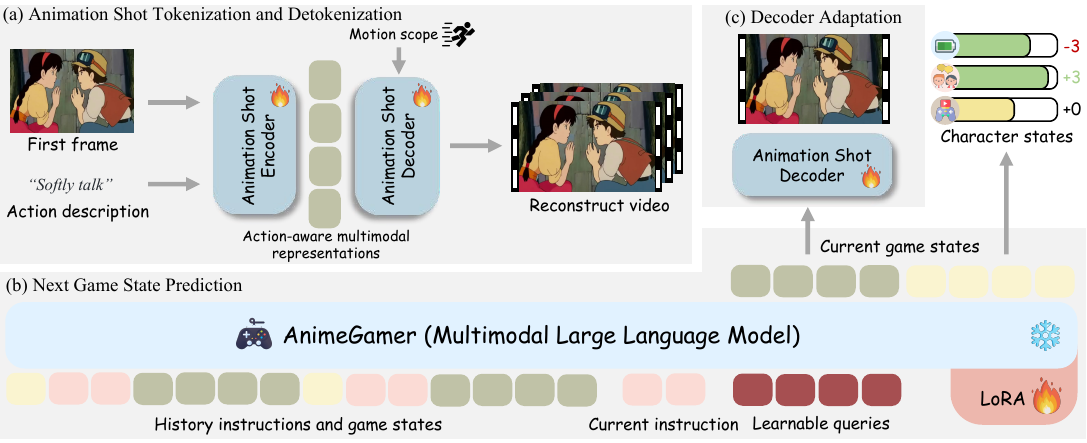}
        \caption{Overview of our AnimeGamer. The training process consists of three phases: (a) We model animation shots using action-aware multimodal representations through an encoder and train a diffusion-based decoder to reconstruct videos, with the additional input of motion scope that indicates action intensity. (b) We train an MLLM to predict the next game state representations by taking the history instructions and game state representations as input. (c) We further enhance the quality of decoded animation shots from the MLLM via an adaptation phase, where the decoder is fine-tuned by taking MLLM's predictions as input.}
	\label{fig: model-mllm}
\end{figure*}

\subsection{Task Formulation}
We focus on the challenging infinite anime life simulation task in this paper. Following prior works~\cite{li2024unbounded,carse1986finite}, we define a round of infinite game as consisting of multiple game states $s$, which serve as feedback to players. Each $s$ is composed of two parts: 1) Dynamic animation shot: an anime clip demonstrating the action of the character; 2) Character state: visualization of a character's stamina, social, and entertainment values to represent their mood and physical health. Models are required to receive open-ended language instructions from players to generate multi-turn game states. 

\subsection{AnimeGamer}

\vspace{1mm} \noindent \textbf{Overview.} The overview of our AnimeGamer is illustrated in Figure~\ref{fig: model-mllm}. We model an animation shot as action-aware multimodal representation by training an animation shot encoder $\mathcal{E}_a$, with an animation shot decoder $\mathcal{D}_a$ based on video diffusion models to decode the representation into high-quality video clips. Next, we introduce an MLLM to predict each game state representation with multimodal input. We further enhance the quality of decoded animation shots from the MLLM via an adaptation phase, where the decoder is fine-tuned by taking MLLM's predictions as input.

\vspace{1mm} \noindent \textbf{Animation Shot Tokenization and Detokenization.} The alignment of a character's visual features and actions in an anime clip with player instructions is crucial for the gaming experience. However, existing MLLM-based methods primarily predict text-only~\cite{xiang2024pandora} or image-only~\cite{zhao2024moviedreamer} representations to align with generative diffusion models. They are limited by the significant loss of visual and motion information in a video clip, resulting in inconsistency in gameplay. To address this, we model an animation shot as action-aware multimodal representation $s_a$ that serve as a bridge for the MLLM and $\mathcal{D}_a$. As illustrated in Figure~\ref{fig: model-mllm}, we decompose an animation shot into the following three parts: 1) Overall visual reference $f_v$, which is captured by CLIP~\cite{radford2021learning} embeddings of the first frame of an anime clip; 2) Action description $f_{md}$, a short motion prompt focusing on the characters' action in the video (e.g., ``Softly talk"), which is represented by T5~\cite{raffel2020exploring} text embeddings; 3) Motion scope $f_{ms}$, we represent the intensity of character's action in a video by optical flow\footnote{Detailed in Appendix A.}. As depicted in Figure~\ref{fig:  model-cogvideo}, the animation shot is encoded by $\mathcal{E}_a$ as follows: 
\begin{equation}
\begin{aligned}
 s_a &=  \mathcal{E}_a(f_{md},f_{v}) \\
 \mathcal{E}_a &= \text{Concat}(\text{LN}(\text{MLP}(x)),\text{LN}(\text{MLP}(y))),
\label{eq: action-aware multimodal representation}
\end{aligned}
\end{equation}
where MLP stands for multi-layer perception for dimension alignment, LN stands for layer normalization to align feature scale, and Concat represents the concatenation operation along the token dimension. Finally, $f_{ms}$ will serve as an additional condition for $\mathcal{D}_a$ to control the motion scope in the output dynamic animation shot.

\begin{figure*}[!t]
	\centering
	\includegraphics[width=\textwidth]{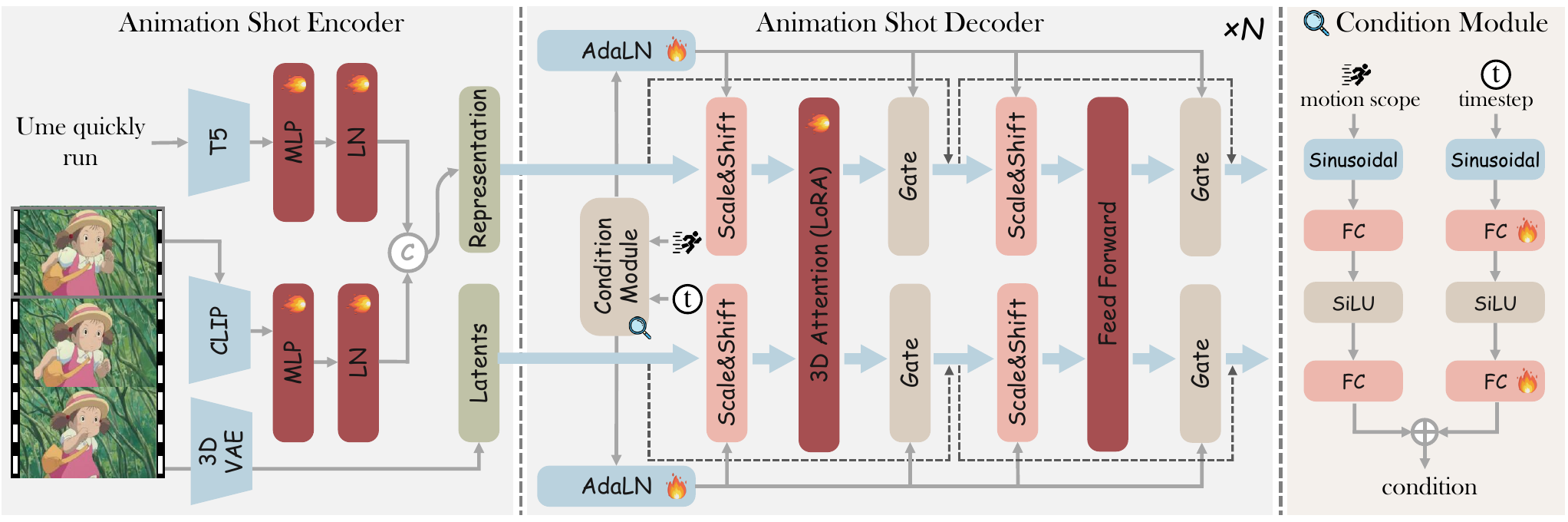}
     \vspace{-5pt}
	\caption{Architecture of animation shot encoder and decoder. The action-aware multimodal representation integrates visual features of the first frame with textual features of action description, and serve as the input to the modulation module of the decoder. Additional motion scope indicating action intensity is injected using a condition module.}
	\label{fig: model-cogvideo}
    \vspace{-10pt}
\end{figure*}

To decode the multimodal representation into high-quality video, we introduce a decoder $\mathcal{D}_a$ upon a video diffusion model CogvideoX~\cite{yang2024cogvideox} by replacing the original text features with the action-aware multimodal representation. In addition, we introduce $f_{ms}$ as an additional generation condition to control action intensity. As illustrated in Figure~\ref{fig: model-cogvideo}, $f_{ms}$ is embedded using sinusoidal functions and several fully-connected (FC) layers activated by SiLU~\cite{hendrycks2016gaussian}, which is then added to the timestep embedding $f_{t}$.

As for training, we first align $s_a$ with the input space of $\mathcal{D}_a$ by optimizing only $\mathcal{E}_a$ as warm-up. We initially encode an input video $x$ into a latent code $z$ using the 3D-Variational Autoencoder from~\cite{yang2024cogvideox}. Next, the noisy latent code $z_t$ at timestep $t$ serves as the input for the denoising DiT $\epsilon_\theta$ with text condition $c$ and $s_a$. The training objective for this process is defined as follows:
\begin{equation}
\label{eq:diffusion loss}
\mathcal{L} = \mathbb{E}_{z,c,s_a,\epsilon \sim \mathcal{N}(0,1),t} \left[ \| \epsilon - \epsilon_\theta (z_t, t, c, s_a) \|_2^2 \right],
\end{equation}
where $\epsilon$ represents random noise sampled from a standard Gaussian distribution. Then, we jointly train $\mathcal{E}_a$ and $\mathcal{D}_a$, and the training loss remains consistent with Equation~\ref{eq:diffusion loss}.

\vspace{1mm} \noindent \textbf{Game State Prediction with MLLM.} Recent advancements in MLLM have demonstrated significant progresses in unified comprehension and generation~\cite{chowdhery2023palm,ge2024divot,yang2024seed,zhu2023vl}. Inspired by this, we utilize MLLM as a ``game engine" to perform infinite anime life simulation by next game state prediction. As shown in Figure~\ref{fig: model-mllm}, AnimeGamer takes multimodal historical context and the current instruction as inputs to generate the next game state. For $s_a$, we employ N learnable queries as input and continuously output N action-aware multimodal representations from the MLLM with full attention. Here, we set N = 226 to align the pre-trained model of $\mathcal{D}_a$ to reduce consumption costs. For $s_{c}$, we predict the three character states, as well as $f_{ms}$ as an additional generation control. We treat these as discrete targets and add special tokens\footnote{See Appendix A for more details.} to format the generation after $s_a$.

During training, we sample a random-length subset from the multi-turn data for each iteration. We initialized AnimeGamer with the weight of Mistral-7B~\cite{jiang2023mistral} and task the model to continuously output the next $s_a$ with MSE loss, and perform next-token prediction to genearte $s_c$ and $f_{ms}$, which as optimized with Cross Entropy loss. The overall training loss is as follows: 
\begin{equation}
\mathcal{L} = \mathcal{L}_{\text{CE}} + \alpha \mathcal{L}_{\text{MSE}},
\label{eq: loss}
\end{equation}
where $\alpha$ is the weight of the loss term $\mathcal{L}_{\text{MSE}}$.

\vspace{1mm} \noindent \textbf{Decoder Adaptation.} The separate training of the MLLM and $\mathcal{D}_a$ conserves memory but risks potential misalignment between the latent spaces of the MLLM output and $\mathcal{D}_a$, which may lead to artifacts in the generated videos. To mitigate this issue, we conduct adaptation training where only $\mathcal{D}_a$ is trained. Conditioned on the output embeddings of the MLLM, $\mathcal{D}_a$ is expected to generate anime shots that are pixel-level aligned with the ground truth.

\vspace{1mm} \noindent \textbf{Inference.} During the inference process, the historical action-aware multimodal representations are projected into the input space of the MLLM using a linear resampler. To enable theoretically infinite generation, we follow previous works~\cite{ren2024timechat} to adopt the sliding window technique for multimodal generation with a train-short-test-long scheme.

\subsection{Dataset Construction}

Training AnimeGamer requires multi-turn character-centric video data with contextual coherence. However, existing anime datasets~\cite{pan2024sakuga,siyao2022animerun,jiang2024exploring, kim2022animeceleb} mainly focus on single scene or are closed-sourced, which limits their application to this challenging task. Noticing that anime films are an ideal data source due to their sufficient time span, narrative coherence and easy accessibility, we construct a pipeline to obtain the required training data from them. Specifically, we collect 10 popular anime films and split them into approximately 20,000 video clips, each containing 16 frames at 480 × 720 resolution. We uniformly sample 4 frames from each video clip as input for InternVL~\cite{chen2024internvl}, prompting it to obtain character movement, background, and character states in the video. Additionally, we collect images of the main characters and prompt InternVL to label them in each frame to ensure character consistency. Players can customize their favorite characters following this pipeline\footnote{See Appendix B for pipeline construction details.}. 

\section{Experiments}

\vspace{-5pt}
\begin{table*}[!t]
  \caption{Quantitative comparison with baseline models on automatic metrics. \textbf{Bold} indicate the best performance.}
  \label{tab: Quantitative comparison-1}
  \centering
\resizebox{1\textwidth}{!}{
\begin{tabular}{cccccccccc}
\toprule
\multicolumn{1}{c}{\multirow{2}{*}{Model}} & \multicolumn{2}{c}{Character Consistency} & \multicolumn{2}{c}{Semantic Consistency} & \multicolumn{2}{c}{Motion Quality} & \multicolumn{2}{c}{State Update}  & \multirow{2}{*}{Inference Time (s/turn) $\downarrow$} \\
\cmidrule(lr){2-3}  \cmidrule(lr){4-5} \cmidrule(lr){6-7} \cmidrule(lr){8-9} 
\multicolumn{1}{c}{}                       & CLIP-I $\uparrow$             & DreamSim $\uparrow$           & CLIP-T $\uparrow$            & $\text{CLIP-T}^{\text{E}}$$\uparrow$              & ACC-F $\uparrow$           & MAE-F $\downarrow$          & ACC-S $\uparrow$          & MAE-S    $\downarrow$       &                                            \\
\midrule
GSC                                        & 0.7862              & 0.5019              & 0.3331              & 0.3142             & 0.3163           & 0.8263          & 0.6773          & 0.5888          & 50                                         \\
GFC                                        & 0.7662              & 0.5797              & 0.3325              & 0.3123             & 0.2923           & 1.0212          & 0.6771          & 0.5888          & 63                                         \\
GC                                         & 0.7960              & 0.6416              & 0.3339              & 0.3158             & 0.4249           & 0.7223          & \textbf{0.6779} & 0.5888          & 25                                         \\
\cellcolor[HTML]{E6F0E8}\textbf{AnimeGamer}                          &\cellcolor[HTML]{E6F0E8}\textbf{0.8132}     &\cellcolor[HTML]{E6F0E8}\textbf{0.7403}     &\cellcolor[HTML]{E6F0E8}\textbf{0.4161}     &\cellcolor[HTML]{E6F0E8}\textbf{0.4012}    &\cellcolor[HTML]{E6F0E8}\textbf{0.6744}  &\cellcolor[HTML]{E6F0E8}\textbf{0.4238} &\cellcolor[HTML]{E6F0E8}0.6773          &\cellcolor[HTML]{E6F0E8}\textbf{0.5872} &\cellcolor[HTML]{E6F0E8}\textbf{24}                               \\
\bottomrule
\end{tabular}
}
\end{table*}

\begin{table*}[!t]
  \caption{Quantitative comparison with baseline models on MLLM judgement and human evaluation. \textbf{Bold} indicate the best performance.}
  \label{tab: Quantitative comparison-2}
  \centering
\resizebox{1\textwidth}{!}{
\begin{tabular}{ccccccccccccc}
\toprule
\multicolumn{1}{c}{\multirow{2}{*}{Model}} & \multicolumn{2}{c}{Overall $\uparrow$ } & \multicolumn{2}{c}{Instruction Following $\uparrow$ } & \multicolumn{2}{c}{Contexual Consistency $\uparrow$ } & \multicolumn{2}{c}{Chracter Consistency $\uparrow$ } & \multicolumn{2}{c}{Style consistency $\uparrow$ } & \multicolumn{2}{c}{State Update $\uparrow$ } \\
\cmidrule(lr){2-3}  \cmidrule(lr){4-5} \cmidrule(lr){6-7} \cmidrule(lr){8-9}  \cmidrule(lr){10-11} \cmidrule(lr){12-13}
\multicolumn{1}{c}{}                       & GPT-4V        & Human       & GPT-4V               & Human              & GPT-4V              & Human               & GPT-4V              & Human              & GPT-4V            & Human             & GPT-4V          & Human          \\
\midrule
GSC                                        & 5.35          & 2.29        & 6.13                 & 2.96               & 5.44                & 2.71                & 5.33                & 2.96               & 5.57              & 5.77              & 8.38            & 9.92           \\
GFC                                        & 4.96          & 4.27        & 5.51                 & 3.57               & 4.73                & 3.20                 & 6.22                & 3.76               & 4.84              & 3.62              & 8.38            & 9.92           \\
GC                                         & 6.42          & 7.38        & 7.29                 & 7.37               & 6.58                & 6.89                & 7.49                & 7.55               & 6.57              & 6.10               & \textbf{8.39}   & \textbf{9.94}  \\
\cellcolor[HTML]{E6F0E8}\textbf{AnimeGamer}                          &\cellcolor[HTML]{E6F0E8}\textbf{8.36} &\cellcolor[HTML]{E6F0E8}\textbf{10.00} &\cellcolor[HTML]{E6F0E8}\textbf{9.14}        &\cellcolor[HTML]{E6F0E8}\textbf{9.95}        &\cellcolor[HTML]{E6F0E8}\textbf{8.41}       &\cellcolor[HTML]{E6F0E8}\textbf{9.95}       &\cellcolor[HTML]{E6F0E8}\textbf{9.11}       &\cellcolor[HTML]{E6F0E8}\textbf{9.86}      &\cellcolor[HTML]{E6F0E8}\textbf{7.52}     &\cellcolor[HTML]{E6F0E8}\textbf{9.95}     &\cellcolor[HTML]{E6F0E8}\textbf{8.39}   &\cellcolor[HTML]{E6F0E8}\textbf{9.94} \\
\bottomrule
\end{tabular}
}
\end{table*}

\subsection{Baselines}
To the best of our knowledge, there is a lack of open-source approaches for this challenging task. For comparison, we tailor related SOTA models to this task. We use Gemini-1.5~\cite{team2024gemini} as a router LLM to comprehend dialogues and generate character states and generation instructions. Based on this, we construct three baseline methods as follows:
\begin{itemize}
    \item  \textbf{GC}: We fine-tune a T2V model CogvideoX to generate animation shot output.
    \item  \textbf{GFC}: We fine-tune a T2I model Flux~\cite{flux2024} and further process the image results using a pre-trained I2V model CogvideoX-I2V to render the final video.
    \item \textbf{GSC}: We integrate CogvideoX-I2V into the story visualization model StoryDiffusion~\cite{zhou2025storydiffusion} as a tuning-free method for comparison.
\end{itemize}
 
We follow the task setting of Unbounded~\cite{li2024unbounded} in infinite game generation, which trains models with custom characters and evaluates them in closed domains. All baselines are trained on the same dataset as our AnimeGamer for fair comparison. See Appendix C for details.

\subsection{Evaluation Benchmark}
To evaluate the quality of infinite game generation, we construct an evaluation benchmark using GPT-4o~\cite{openai-gpt4o}. We randomly select characters from our training data and prompt GPT-4o to simulate multiple infinite games, with each game containing 10 rounds of instructions. We prompt GPT-4o to provide instructions that include characters, movement descriptions, and the environment, along with the corresponding ground-truth character states for each turn. The benchmark comprises 2,000 rounds, featuring 20 characters, 940 distinct movements, and 133 unique environments. See Appendix D for details.

\subsection{Metrics}
We use automatic metrics CLIP-I~\cite{radford2021learning}, DINO-I~\cite{caron2021emerging} and DreamSim~\cite{fu2023dreamsim} to evaluate character consistency by mapping the detected generated characters to the ground-truth, in line with prior works~\cite{li2024unbounded,zhao2024moviedreamer,cheng2024theatergen}. For semantic consistency, we employ CLIP to calculate cosine similarity between the generated video and the input text prompt and environment prompt, denoted as $\text{CLIP-T}$ and $\text{CLIP-T}^{\text{E}}$, respectively. We further utilize an optical flow detection model~\cite{dong2024memflow} to detect the action intensity of the generated video and calculate Mean Absolute Error (MAE) and Accuracy (ACC) with the ground truth motion scope, denoted as MAE-F and ACC-F, respectively. To assess updates in character states, we report both MAE and ACC, denoted as MAE-S and ACC-S, respectively. Furthermore, some researches~\cite{zheng2023judging,li2024unbounded,yang2024seed} employ a more advanced MLLM as judges to assess the outputs of different models. In this study, we utilize GPT-4v as the evaluation MLLM to score the models from various aspects. We also conduct user studies, adhering to previous game generation works~\cite{anjum2024ink,guzdial2021conceptual,hu2024survey}. Please refer to Appendix D and E for more details.

\subsection{Quantitative Comparisons}
The performance comparison results based on automatic metrics, MLLM judgement and human evaluation are presented in Table~\ref{tab: Quantitative comparison-1} and Table~\ref{tab: Quantitative comparison-2}. AnimeGamer outperforms all baseline models in terms of character consistency, semantic consistency, and motion control within the generated animation shots. This can be attributed to the action-aware multimodal representation of animation shots, which enhances controllability and generalizability. Additionally, AnimeGamer performs favorably in contextual consistency and style consistency, due to the multimodal comprehension and generation capabilities of MLLM. In contrast, other baseline models only consider text context, resulting in a decline across all metrics. When it comes to character state updates, AnimeGamer performs similarly to Gemini-1.5. However, using the API of an LLM incurs additional time costs, giving AnimeGamer an advantage in inference time.

\subsection{Qualitative Comparisons}
\begin{figure*}[!t]
  \centering
  \setlength{\tabcolsep}{1.3pt} 
  \begin{tabular}{ccccccccc}
  \multicolumn{8}{c}{\makebox[0pt][c]{\small \hspace{7em} \raisebox{-0.2ex}{\includegraphics[height=1em]{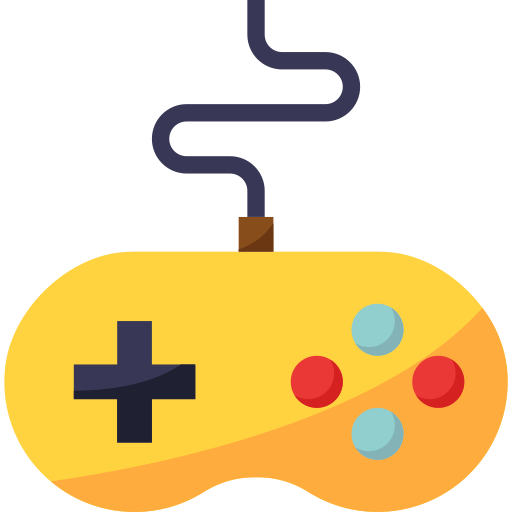}} \hspace{0.2em} Round 1 \hspace{1em} \raisebox{-0.3ex}{\includegraphics[height=1em]{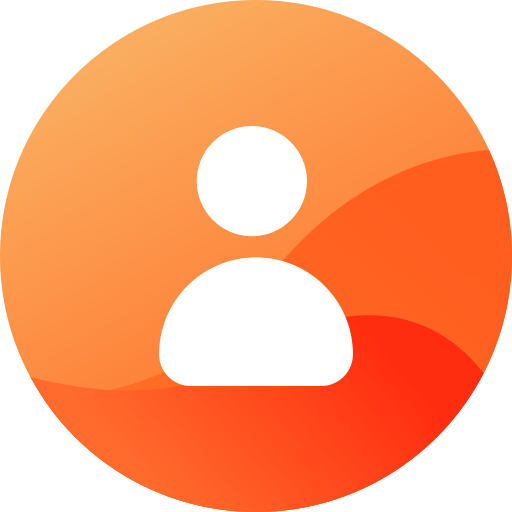}} \hspace{0.2em} \textcolor[HTML]{EFA202}{Pazu} and \textcolor[HTML]{FF4967}{Kiki} \hspace{1em} \raisebox{-0.2ex}{\includegraphics[height=1em]{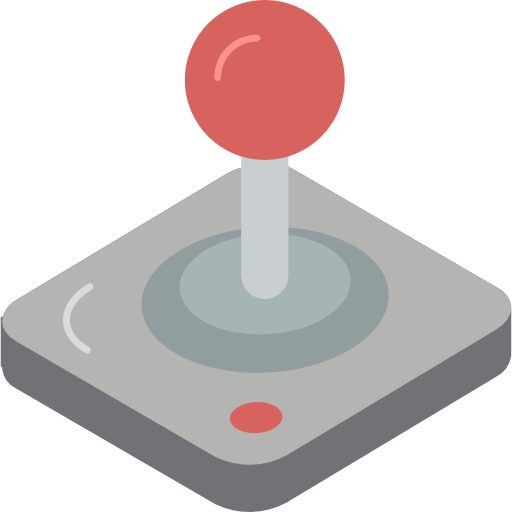}} \hspace{0.2em} Peacefully stand together \hspace{1em} \raisebox{-0.3ex}{\includegraphics[height=1em]{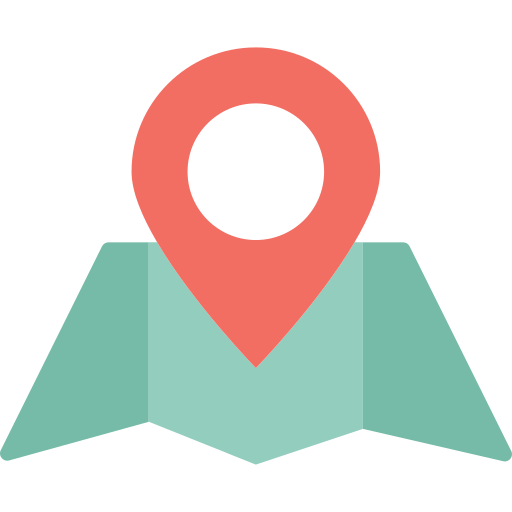}} \hspace{0.2em} Room}} \\
     \raisebox{1.6\height}{\rotatebox[origin=c]{90}{\fontsize{7}{10}\selectfont \textbf{Ours}}} &
    \includegraphics[width=.11\textwidth]{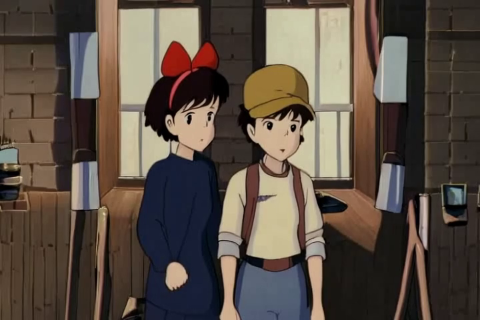} &
    \includegraphics[width=.11\textwidth]{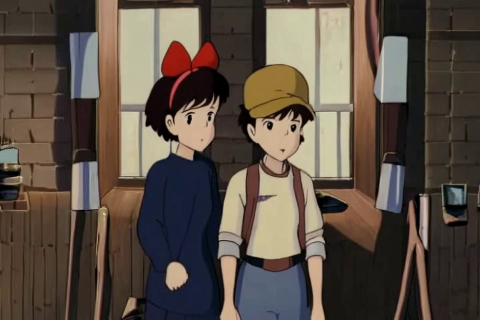} &
    \includegraphics[width=.11\textwidth]{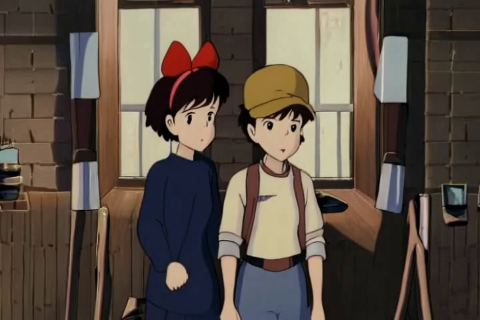} &
    \includegraphics[width=.11\textwidth]{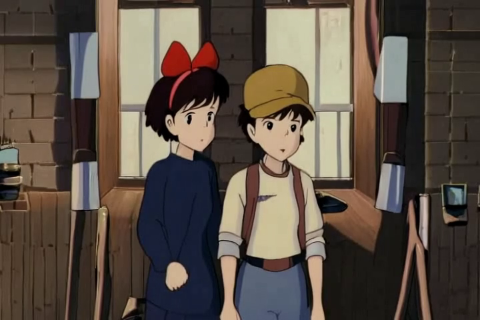} &
    \includegraphics[width=.11\textwidth]{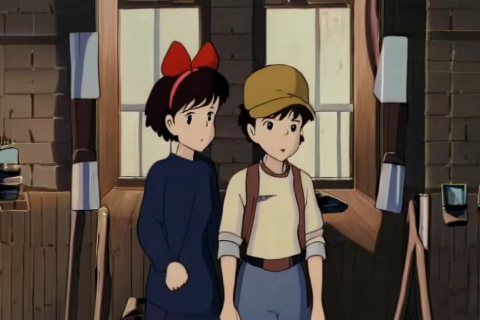} &
    \includegraphics[width=.11\textwidth]{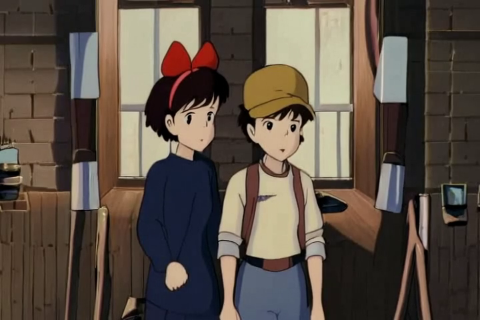} &
    \includegraphics[width=.11\textwidth]{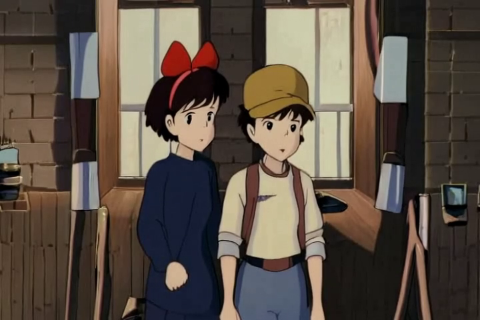} &
    \includegraphics[width=.11\textwidth]{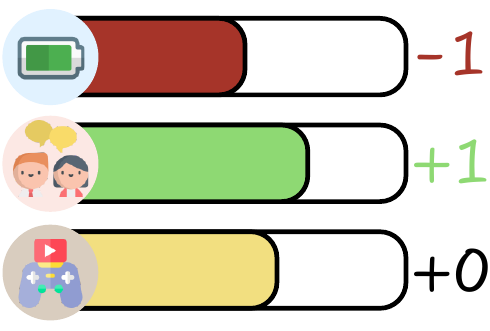}\\
  \multicolumn{4}{c}{\vspace{-14.8pt}} \\
    \raisebox{2\height}{\rotatebox[origin=c]{90}{\fontsize{7}{10}\selectfont GC}} &
    \includegraphics[width=.11\textwidth]{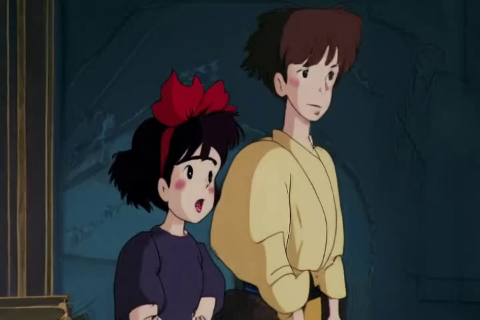} &
    \includegraphics[width=.11\textwidth]{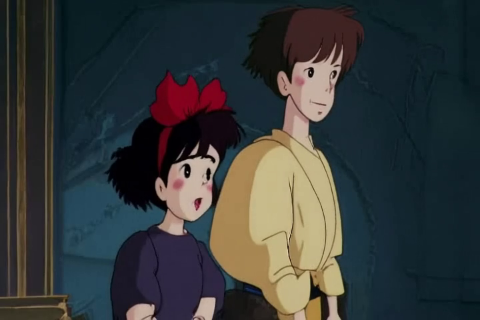} &
    \includegraphics[width=.11\textwidth]{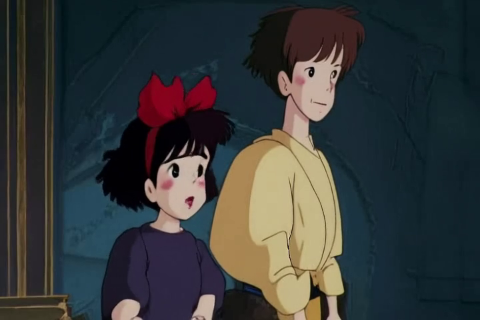} &
    \includegraphics[width=.11\textwidth]{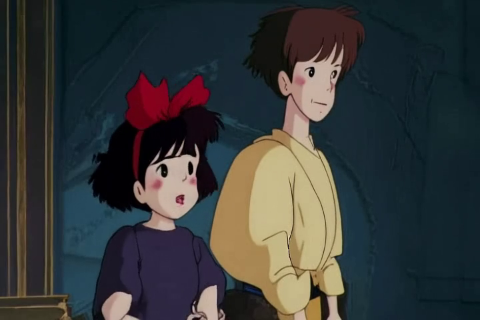} &
    \includegraphics[width=.11\textwidth]{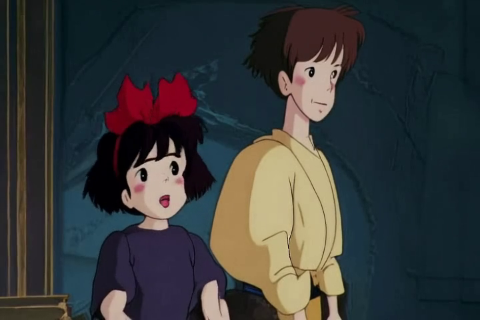} &
    \includegraphics[width=.11\textwidth]{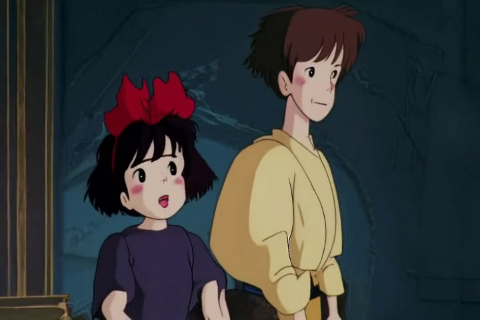} &
    \includegraphics[width=.11\textwidth]{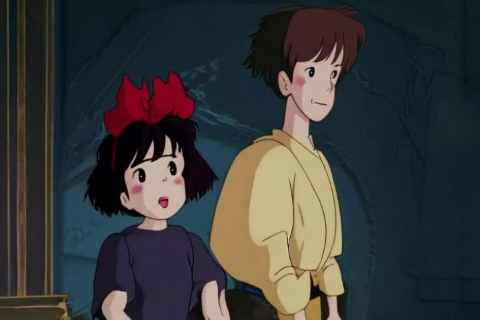} &
    \includegraphics[width=.11\textwidth]{images/main-figure-state/1.pdf}\\
  \multicolumn{4}{c}{\vspace{-14.8pt}} \\
    \raisebox{1.7\height}{\rotatebox[origin=c]{90}{\fontsize{7}{10}\selectfont GFC}} &
    \includegraphics[width=.11\textwidth]{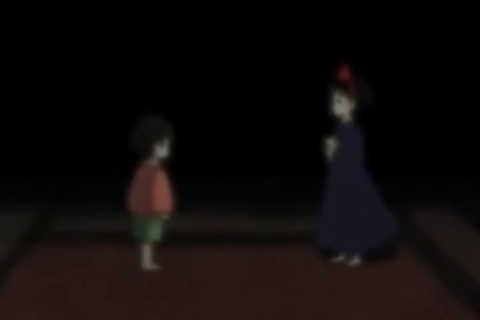} &
    \includegraphics[width=.11\textwidth]{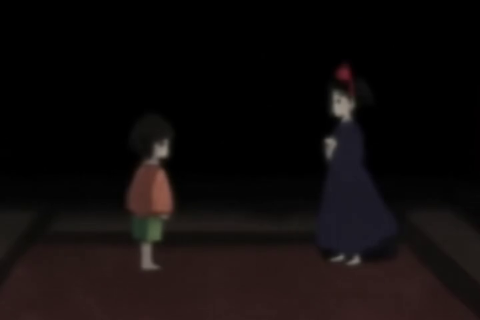} &
    \includegraphics[width=.11\textwidth]{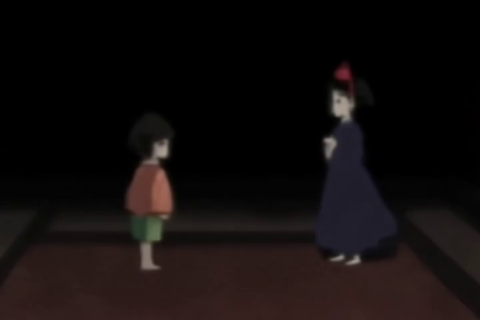} &
    \includegraphics[width=.11\textwidth]{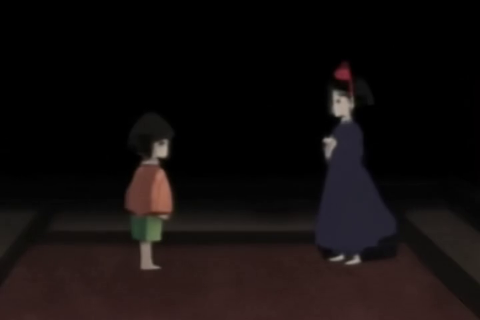} &
    \includegraphics[width=.11\textwidth]{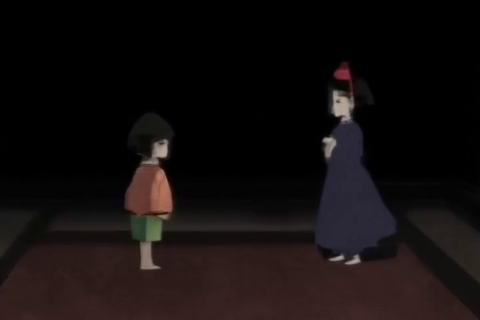} &
    \includegraphics[width=.11\textwidth]{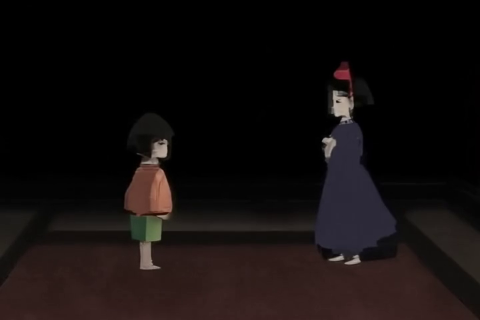} &
    \includegraphics[width=.11\textwidth]{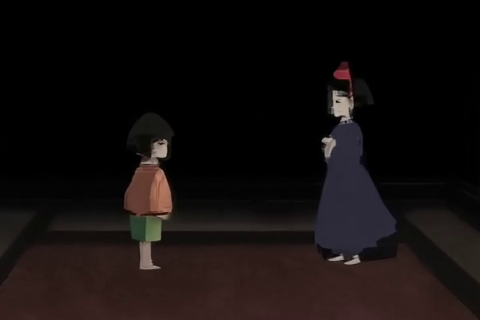} &
    \includegraphics[width=.11\textwidth]{images/main-figure-state/1.pdf}\\
  \multicolumn{4}{c}{\vspace{-14.8pt}} \\
  \multicolumn{8}{c}{\makebox[0pt][c]{\small \hspace{7em} \raisebox{-0.2ex}{\includegraphics[height=1em]{images/icon/game.png}} \hspace{0.2em} Round 2 \hspace{1em} 
  \raisebox{-0.3ex}{\includegraphics[height=1em]{images/icon/profile.png}} \hspace{0.2em} \textcolor[HTML]{EFA202}{Pazu} and \textcolor[HTML]{FF4967}{Kiki} \hspace{1em} \raisebox{-0.2ex}{\includegraphics[height=1em]{images/icon/joystick.png}} \hspace{0.2em} Quietly research the book \hspace{1em} \raisebox{-0.3ex}{\includegraphics[height=1em]{images/icon/map.png}} \hspace{0.2em} Room}} \\
     \raisebox{1.6\height}{\rotatebox[origin=c]{90}{\fontsize{7}{10}\selectfont \textbf{Ours}}} &
    \includegraphics[width=.11\textwidth]{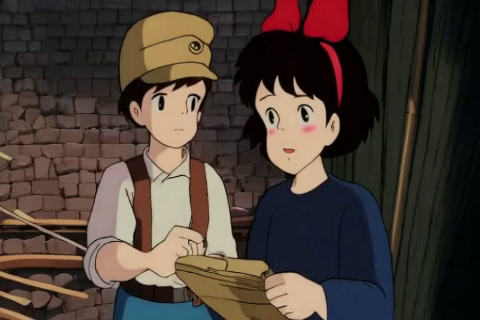} &
    \includegraphics[width=.11\textwidth]{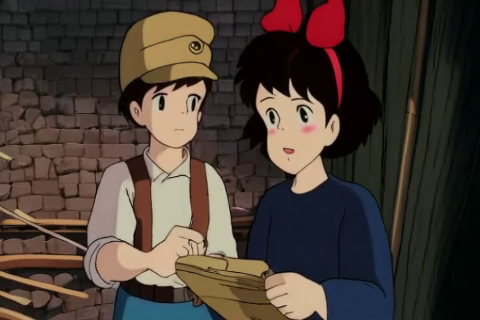} &
    \includegraphics[width=.11\textwidth]{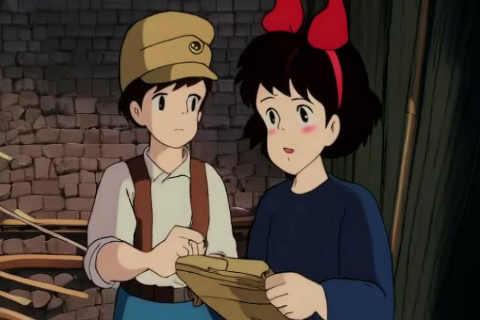} &
    \includegraphics[width=.11\textwidth]{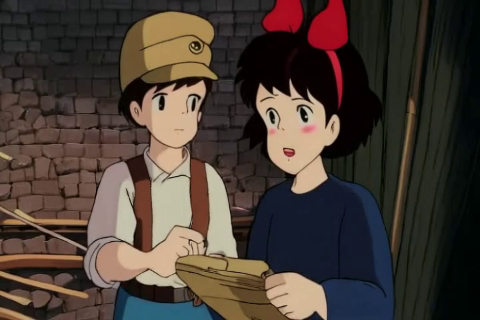} &
    \includegraphics[width=.11\textwidth]{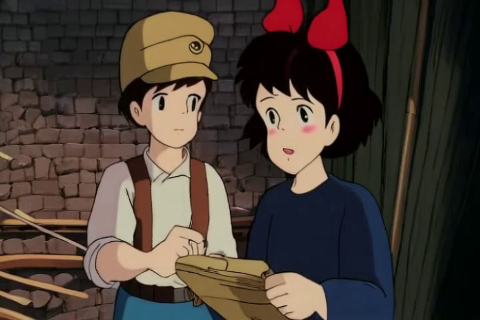} &
    \includegraphics[width=.11\textwidth]{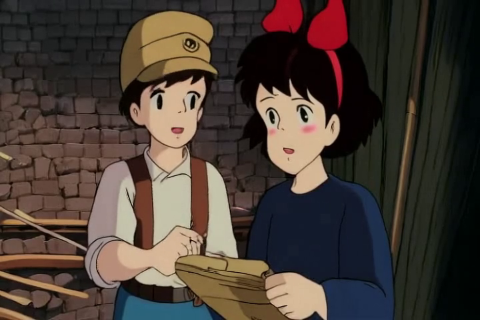} &
    \includegraphics[width=.11\textwidth]{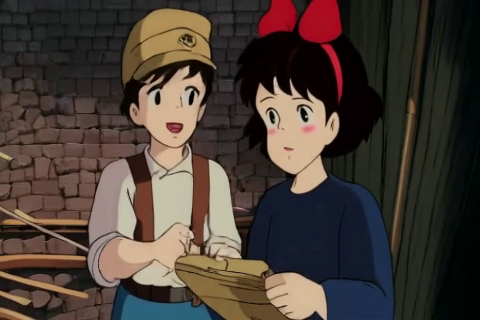} &
    \includegraphics[width=.11\textwidth]{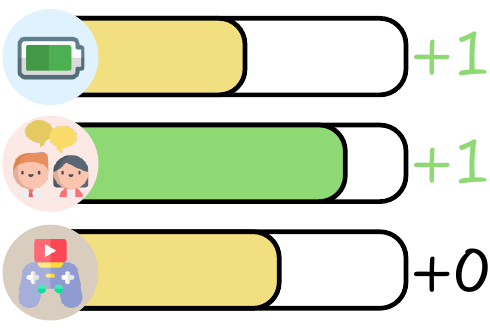}\\
  \multicolumn{4}{c}{\vspace{-14.8pt}} \\
    \raisebox{2\height}{\rotatebox[origin=c]{90}{\fontsize{7}{10}\selectfont GC}} &
    \includegraphics[width=.11\textwidth]{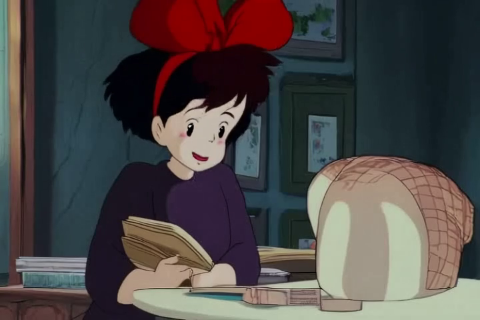} &
    \includegraphics[width=.11\textwidth]{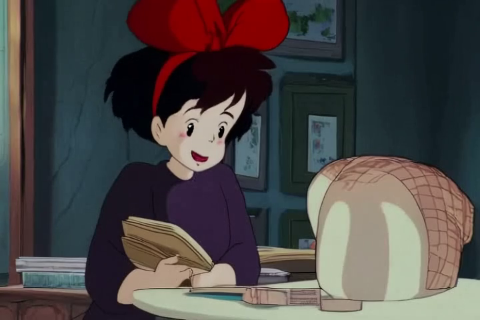} &
    \includegraphics[width=.11\textwidth]{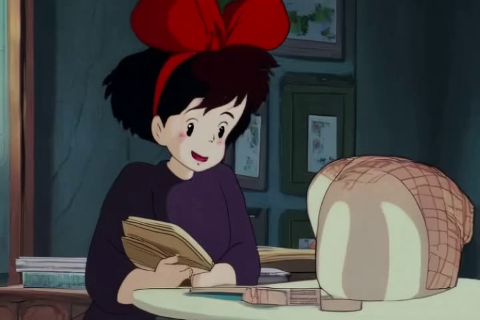} &
    \includegraphics[width=.11\textwidth]{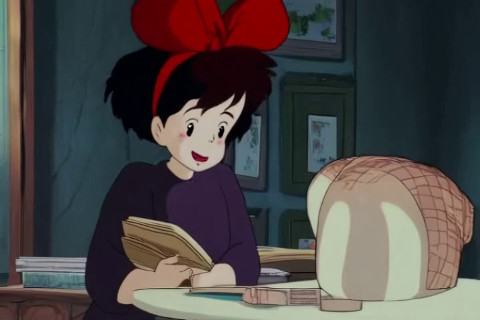} &
    \includegraphics[width=.11\textwidth]{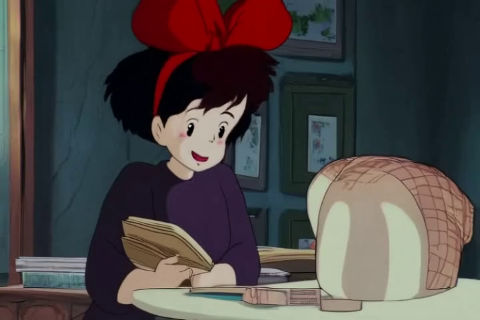} &
    \includegraphics[width=.11\textwidth]{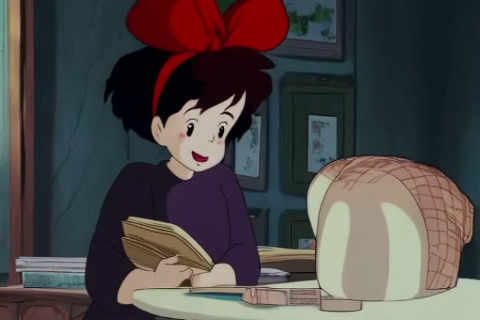} &
    \includegraphics[width=.11\textwidth]{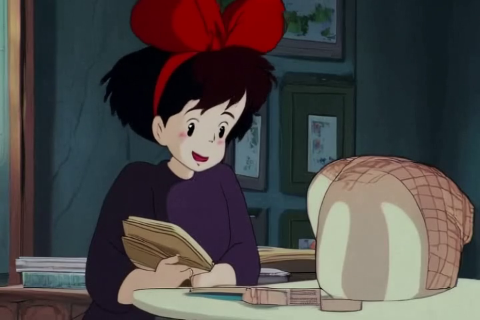} &
    \includegraphics[width=.11\textwidth]{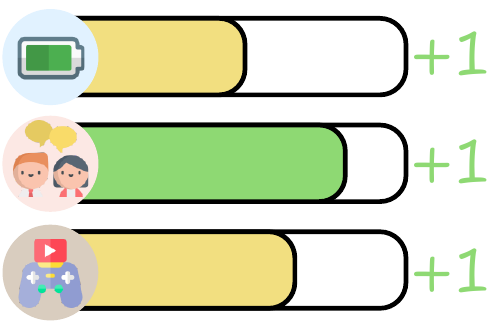}\\
  \multicolumn{4}{c}{\vspace{-14.8pt}} \\
    \raisebox{1.7\height}{\rotatebox[origin=c]{90}{\fontsize{7}{10}\selectfont GFC}} &
    \includegraphics[width=.11\textwidth]{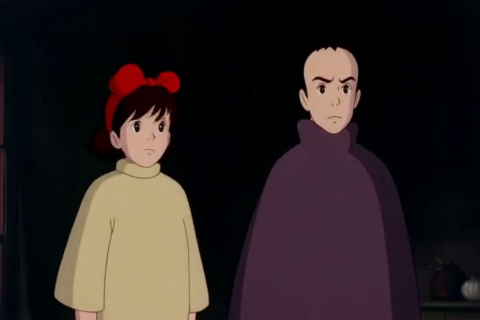} &
    \includegraphics[width=.11\textwidth]{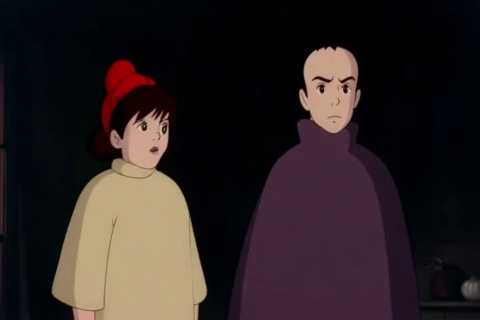} &
    \includegraphics[width=.11\textwidth]{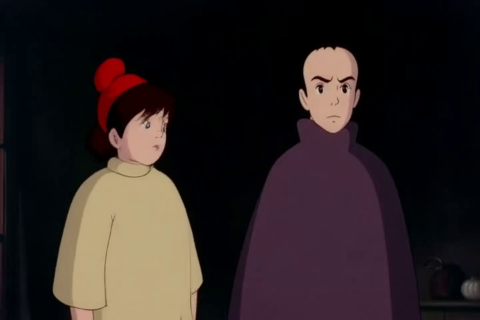} &
    \includegraphics[width=.11\textwidth]{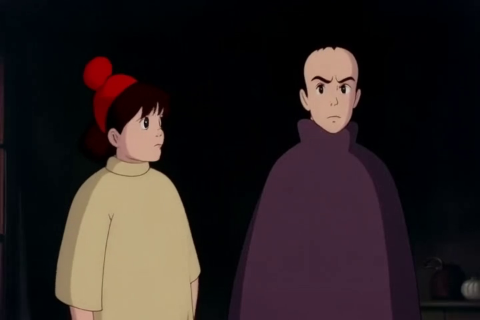} &
    \includegraphics[width=.11\textwidth]{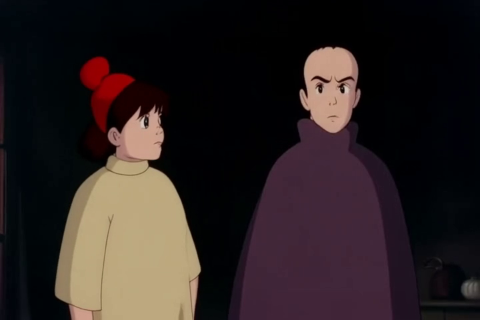} &
    \includegraphics[width=.11\textwidth]{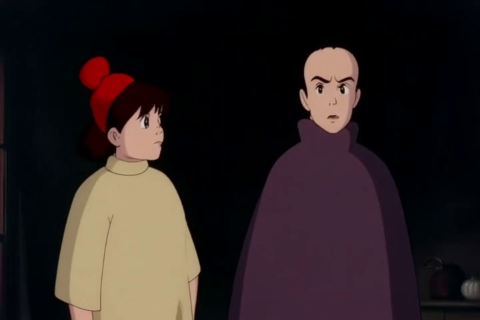} &
    \includegraphics[width=.11\textwidth]{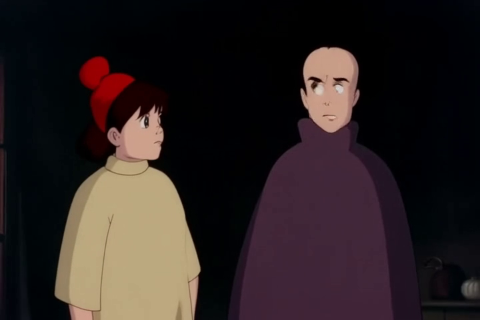} &
    \includegraphics[width=.11\textwidth]{images/main-figure-state/2-1.pdf}\\
  \multicolumn{4}{c}{\vspace{-14.8pt}} \\
  \multicolumn{8}{c}{\makebox[0pt][c]{\small \hspace{7em} \raisebox{-0.2ex}{\includegraphics[height=1em]{images/icon/game.png}} \hspace{0.2em} Round 3 \hspace{1em} \raisebox{-0.3ex}{\includegraphics[height=1em]{images/icon/profile.png}} \hspace{0.2em} \textcolor[HTML]{EFA202}{Pazu} \hspace{1em} \raisebox{-0.2ex}{\includegraphics[height=1em]{images/icon/joystick.png}} \hspace{0.2em} Steadily fly on broomstick \hspace{1em} \raisebox{-0.3ex}{\includegraphics[height=1em]{images/icon/map.png}} \hspace{0.2em} Meadow}} \\
     \raisebox{1.6\height}{\rotatebox[origin=c]{90}{\fontsize{7}{10}\selectfont \textbf{Ours}}} &
    \includegraphics[width=.11\textwidth]{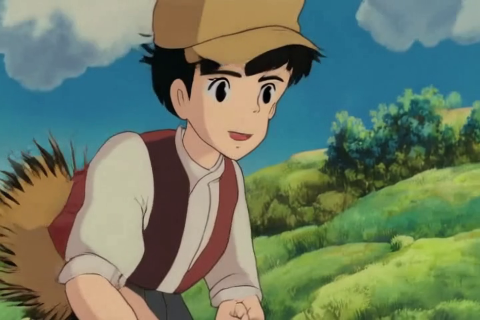} &
    \includegraphics[width=.11\textwidth]{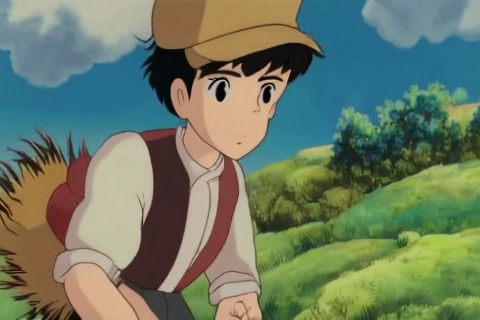} &
    \includegraphics[width=.11\textwidth]{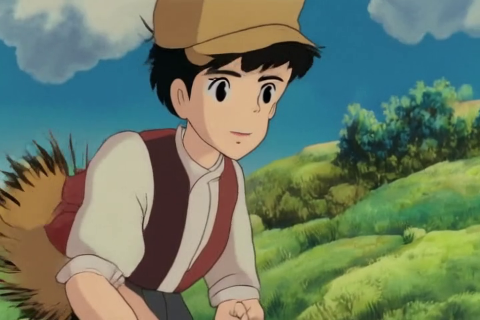} &
    \includegraphics[width=.11\textwidth]{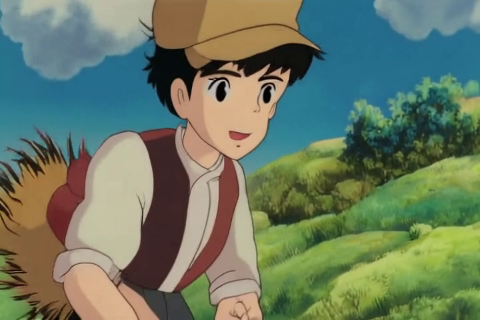} &
    \includegraphics[width=.11\textwidth]{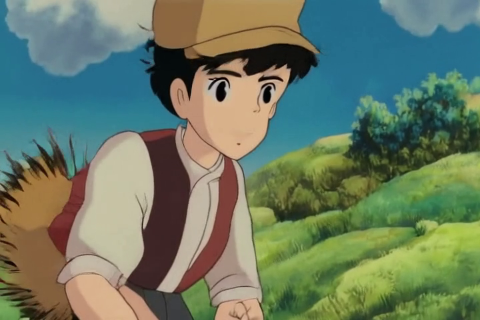} &
    \includegraphics[width=.11\textwidth]{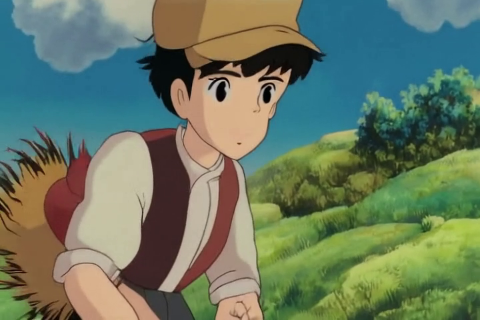} &
    \includegraphics[width=.11\textwidth]{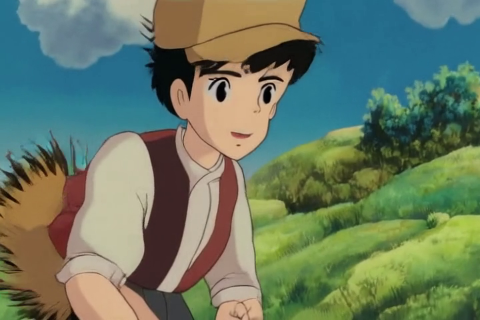} &
    \includegraphics[width=.11\textwidth]{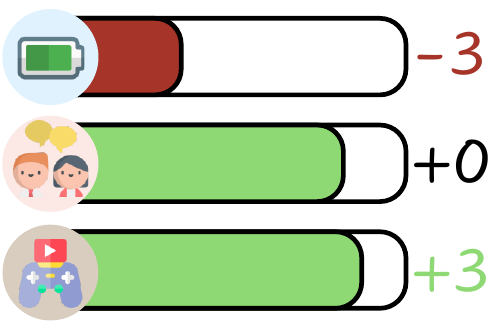}\\
  \multicolumn{4}{c}{\vspace{-14.8pt}} \\
    \raisebox{2\height}{\rotatebox[origin=c]{90}{\fontsize{7}{10}\selectfont GC}} &
    \includegraphics[width=.11\textwidth]{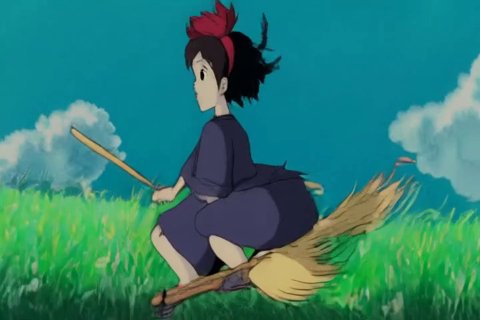} &
    \includegraphics[width=.11\textwidth]{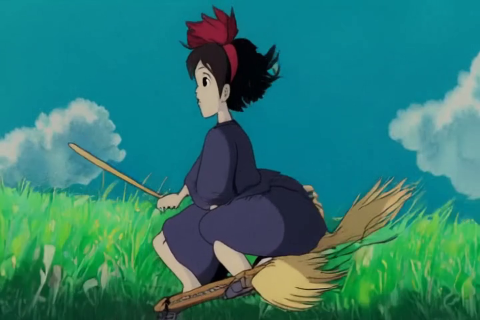} &
    \includegraphics[width=.11\textwidth]{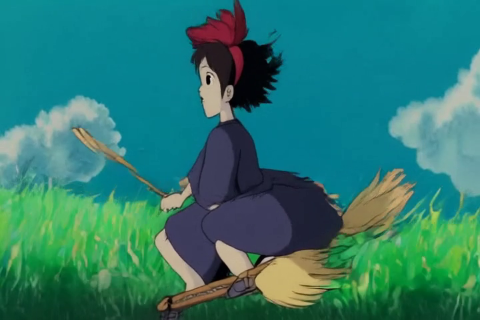} &
    \includegraphics[width=.11\textwidth]{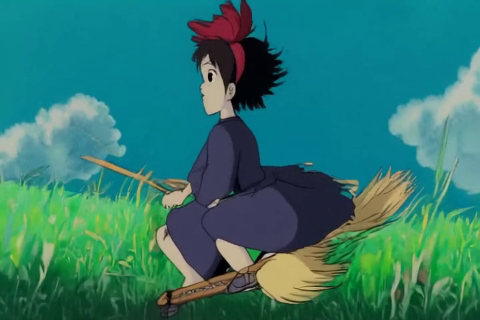} &
    \includegraphics[width=.11\textwidth]{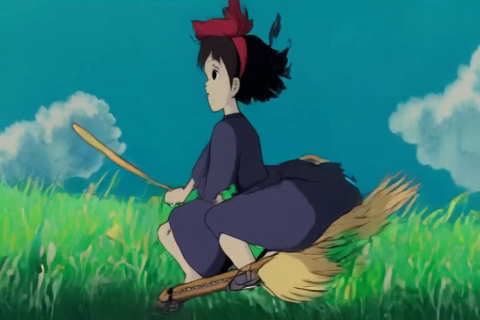} &
    \includegraphics[width=.11\textwidth]{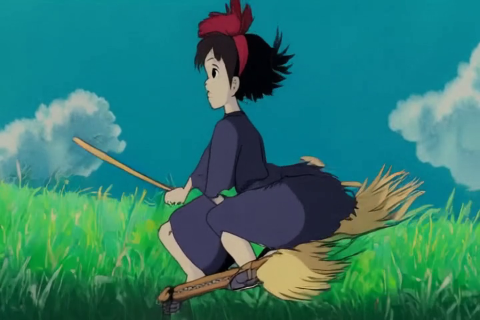} &
    \includegraphics[width=.11\textwidth]{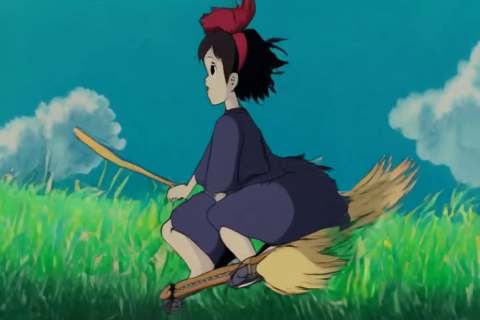} &
    \includegraphics[width=.11\textwidth]{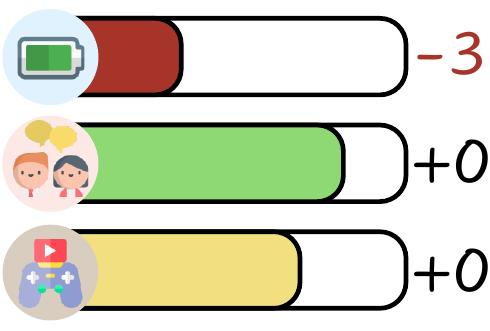}\\
  \multicolumn{4}{c}{\vspace{-14.8pt}} \\
    \raisebox{1.7\height}{\rotatebox[origin=c]{90}{\fontsize{7}{10}\selectfont GFC}} &
    \includegraphics[width=.11\textwidth]{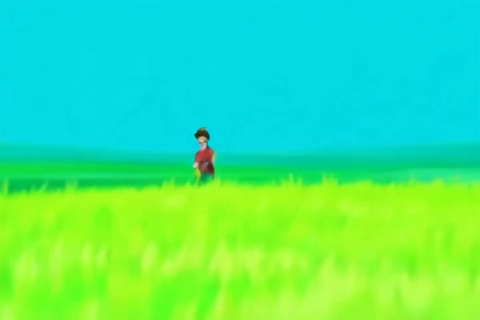} &
    \includegraphics[width=.11\textwidth]{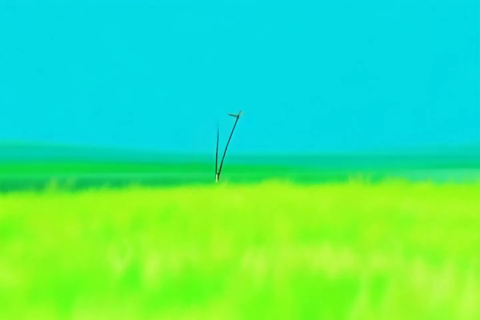} &
    \includegraphics[width=.11\textwidth]{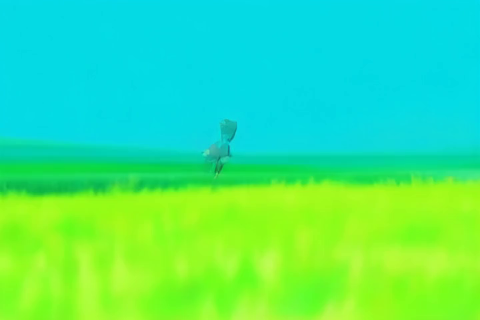} &
    \includegraphics[width=.11\textwidth]{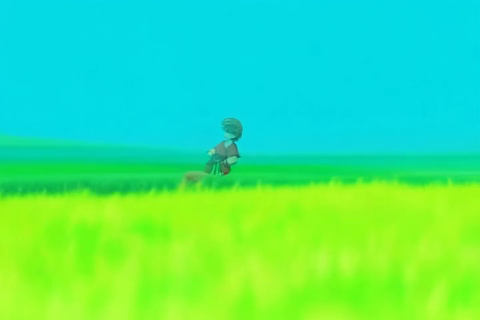} &
    \includegraphics[width=.11\textwidth]{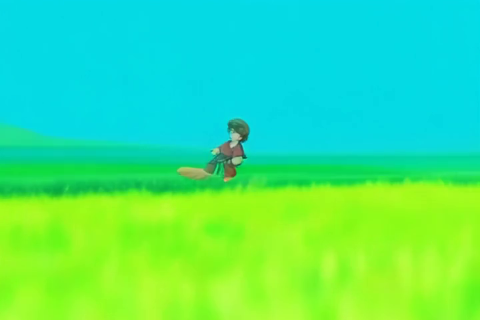} &
    \includegraphics[width=.11\textwidth]{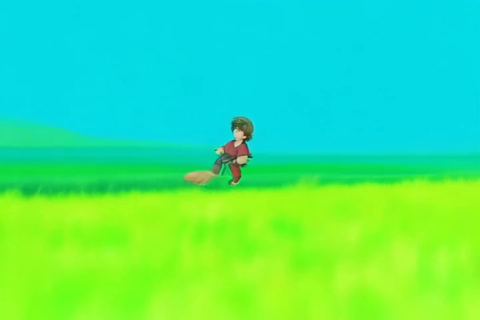} &
    \includegraphics[width=.11\textwidth]{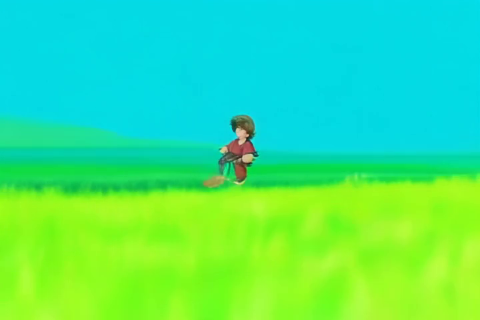} &
    \includegraphics[width=.11\textwidth]{images/main-figure-state/3-1.pdf}\\
  \multicolumn{4}{c}{\vspace{-14.8pt}} \\
  \multicolumn{8}{c}{\makebox[0pt][c]{\small \hspace{7em} \raisebox{-0.2ex}{\includegraphics[height=1em]{images/icon/game.png}} \hspace{0.2em} Round 4 \hspace{1em} \raisebox{-0.3ex}{\includegraphics[height=1em]{images/icon/profile.png}} \hspace{0.2em} \textcolor[HTML]{EFA202}{Pazu} \hspace{1em} \raisebox{-0.2ex}{\includegraphics[height=1em]{images/icon/joystick.png}} \hspace{0.2em} Carefully read map \hspace{1em} \raisebox{-0.3ex}{\includegraphics[height=1em]{images/icon/map.png}} \hspace{0.2em} Cave}} \\
     \raisebox{1.6\height}{\rotatebox[origin=c]{90}{\fontsize{7}{10}\selectfont \textbf{Ours}}} &
    \includegraphics[width=.11\textwidth]{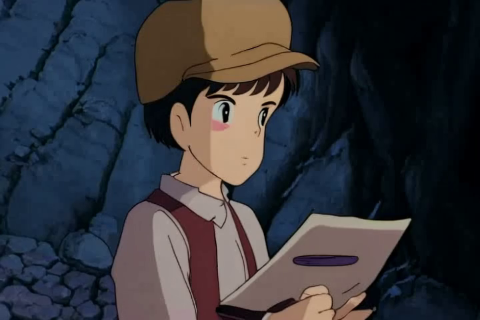} &
    \includegraphics[width=.11\textwidth]{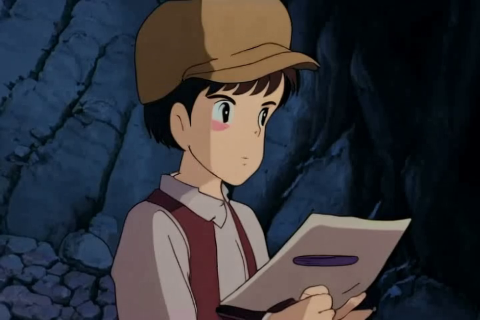} &
    \includegraphics[width=.11\textwidth]{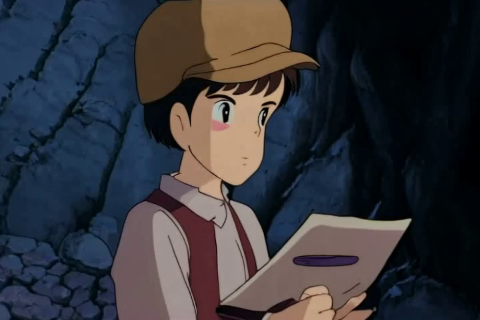} &
    \includegraphics[width=.11\textwidth]{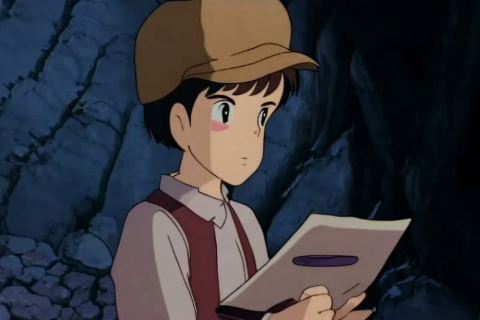} &
    \includegraphics[width=.11\textwidth]{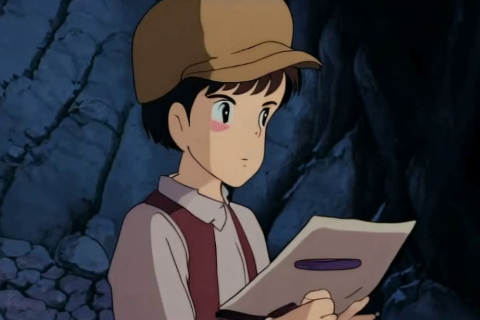} &
    \includegraphics[width=.11\textwidth]{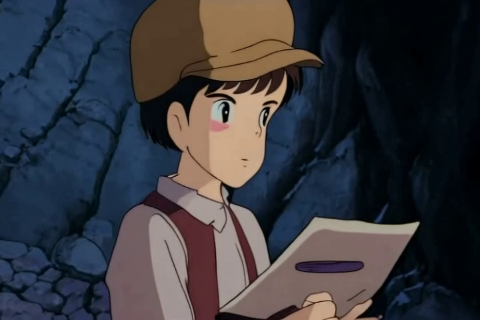} &
    \includegraphics[width=.11\textwidth]{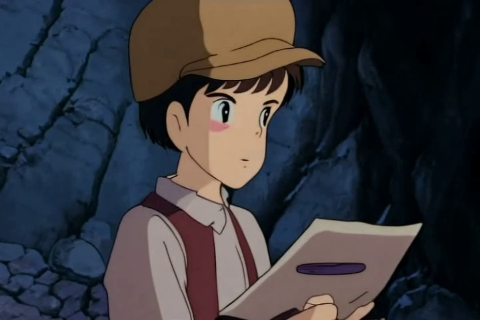} &
    \includegraphics[width=.11\textwidth]{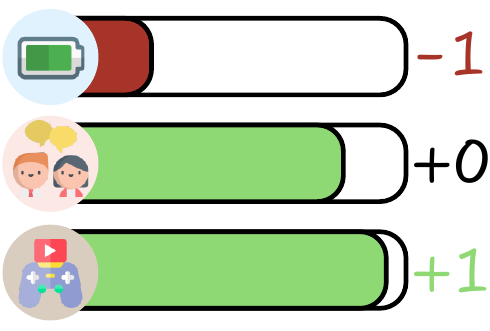}\\
  \multicolumn{4}{c}{\vspace{-14.8pt}} \\
    \raisebox{2\height}{\rotatebox[origin=c]{90}{\fontsize{7}{10}\selectfont GC}} &
    \includegraphics[width=.11\textwidth]{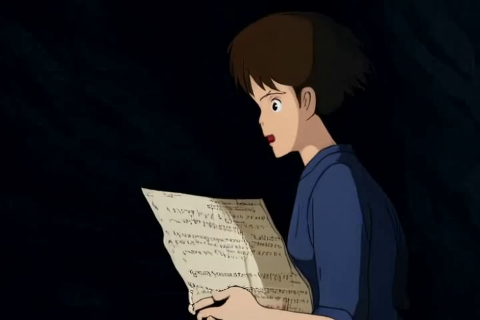} &
    \includegraphics[width=.11\textwidth]{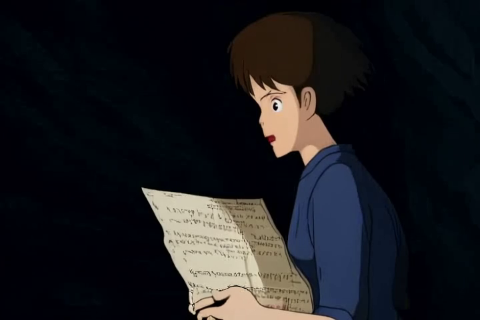} &
    \includegraphics[width=.11\textwidth]{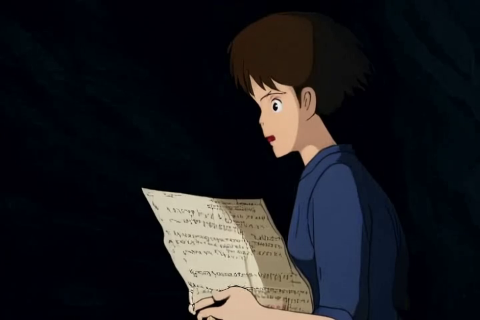} &
    \includegraphics[width=.11\textwidth]{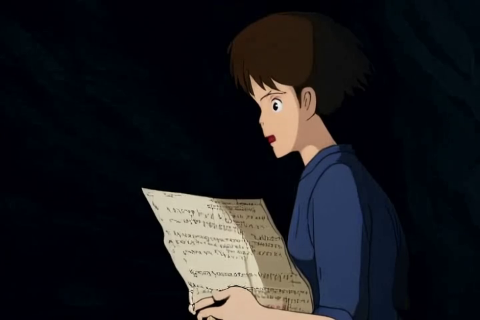} &
    \includegraphics[width=.11\textwidth]{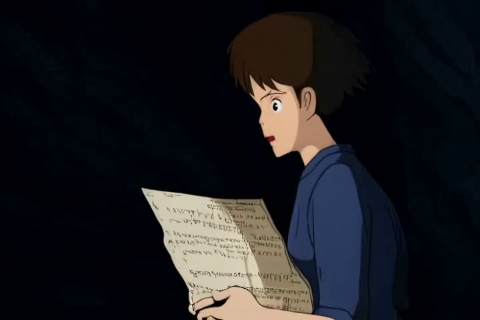} &
    \includegraphics[width=.11\textwidth]{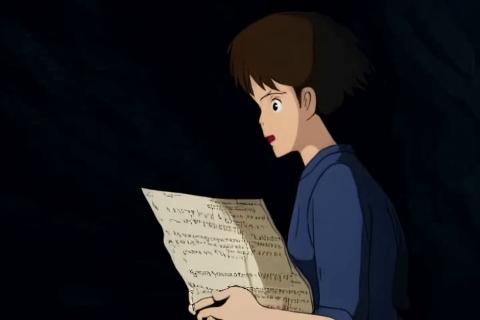} &
    \includegraphics[width=.11\textwidth]{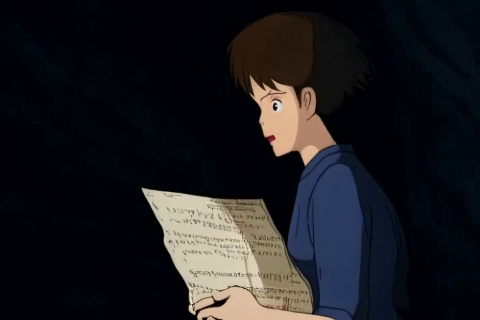} &
    \includegraphics[width=.11\textwidth]{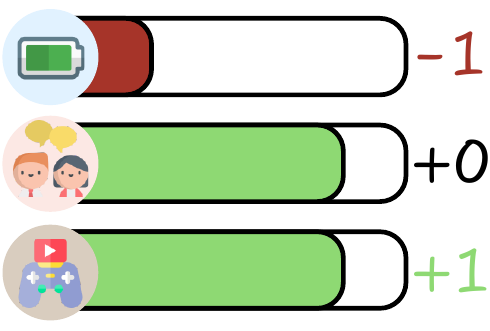}\\
  \multicolumn{4}{c}{\vspace{-14.8pt}} \\
    \raisebox{1.7\height}{\rotatebox[origin=c]{90}{\fontsize{7}{10}\selectfont GFC}} &
    \includegraphics[width=.11\textwidth]{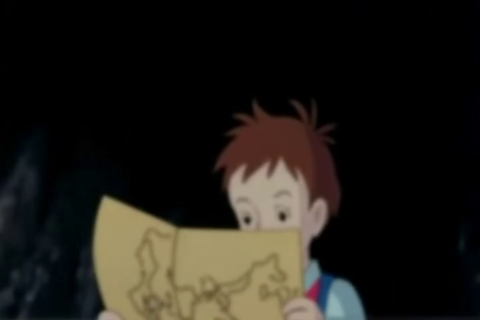} &
    \includegraphics[width=.11\textwidth]{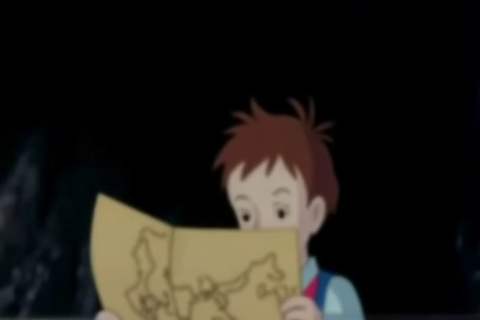} &
    \includegraphics[width=.11\textwidth]{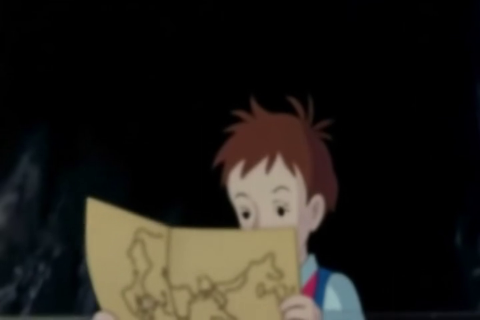} &
    \includegraphics[width=.11\textwidth]{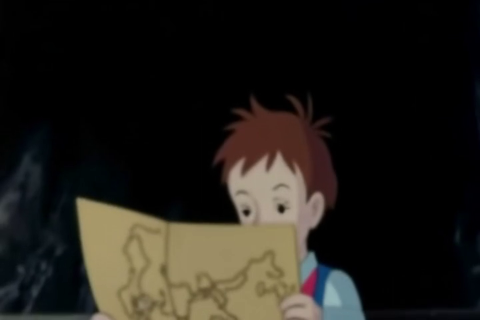} &
    \includegraphics[width=.11\textwidth]{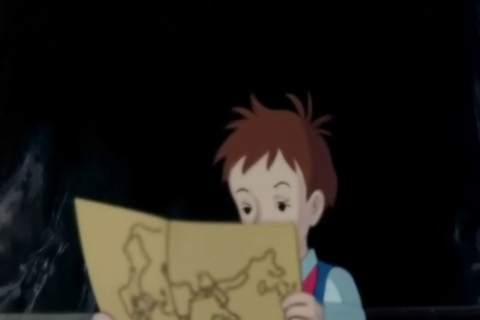} &
    \includegraphics[width=.11\textwidth]{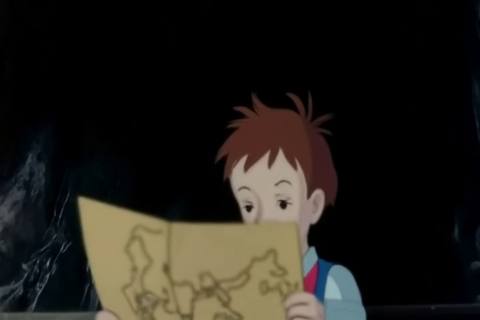} &
    \includegraphics[width=.11\textwidth]{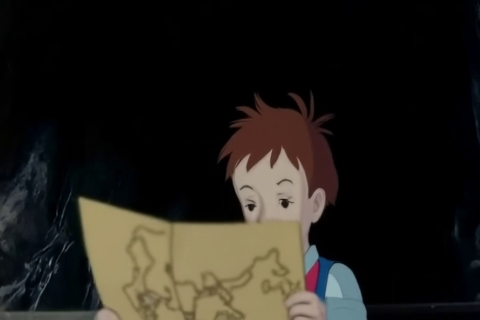} &
    \includegraphics[width=.11\textwidth]{images/main-figure-state/4-1.pdf}\\
  \multicolumn{4}{c}{\vspace{-14.8pt}} \\
  \multicolumn{8}{c}{\makebox[0pt][c]{\small \hspace{7em} \raisebox{-0.2ex}{\includegraphics[height=1em]{images/icon/game.png}} \hspace{0.2em} Round 5 \hspace{1em} \raisebox{-0.3ex}{\includegraphics[height=1em]{images/icon/profile.png}} \hspace{0.2em} \textcolor[HTML]{EFA202}{Pazu} and \textcolor[HTML]{657AA5}{Sheeta} \hspace{1em} \raisebox{-0.2ex}{\includegraphics[height=1em]{images/icon/joystick.png}} \hspace{0.2em} Gently talk to each other \hspace{1em} \raisebox{-0.3ex}{\includegraphics[height=1em]{images/icon/map.png}} \hspace{0.2em} Cave}} \\
     \raisebox{1.6\height}{\rotatebox[origin=c]{90}{\fontsize{7}{10}\selectfont \textbf{Ours}}} &
    \includegraphics[width=.11\textwidth]{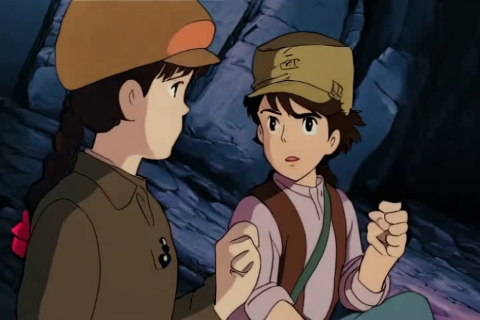} &
    \includegraphics[width=.11\textwidth]{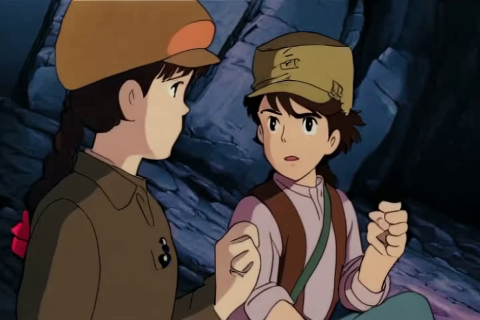} &
    \includegraphics[width=.11\textwidth]{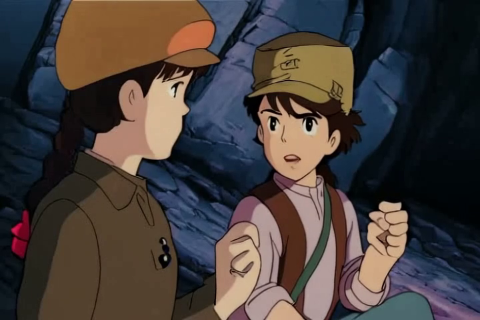} &
    \includegraphics[width=.11\textwidth]{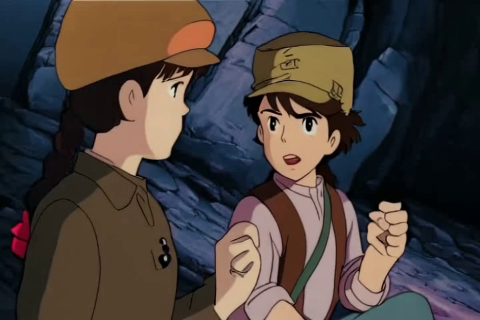} &
    \includegraphics[width=.11\textwidth]{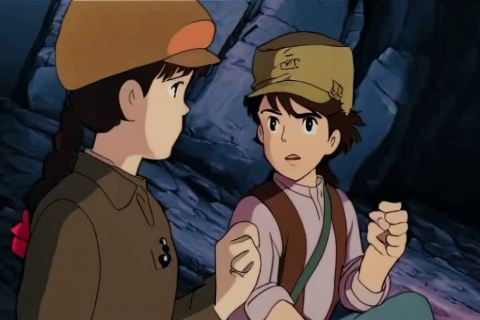} &
    \includegraphics[width=.11\textwidth]{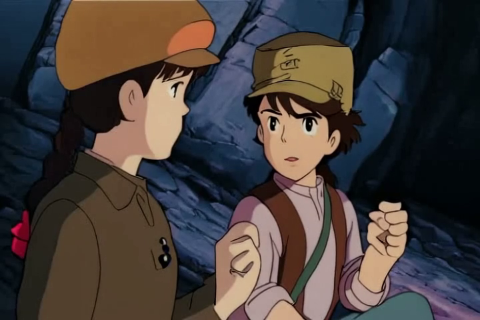} &
    \includegraphics[width=.11\textwidth]{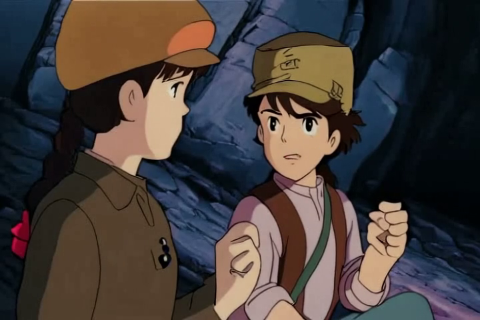} &
    \includegraphics[width=.11\textwidth]{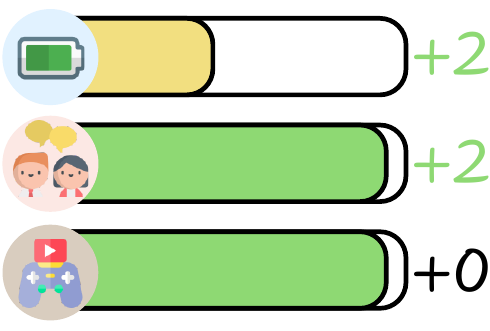}\\
  \multicolumn{4}{c}{\vspace{-14.8pt}} \\
    \raisebox{2\height}{\rotatebox[origin=c]{90}{\fontsize{7}{10}\selectfont GC}} &
    \includegraphics[width=.11\textwidth]{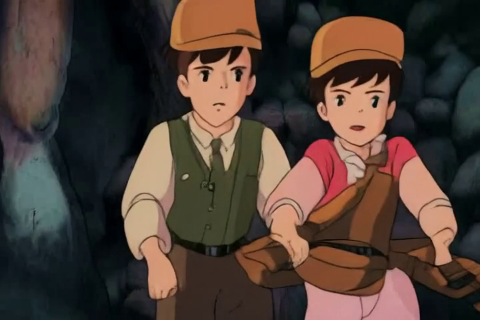} &
    \includegraphics[width=.11\textwidth]{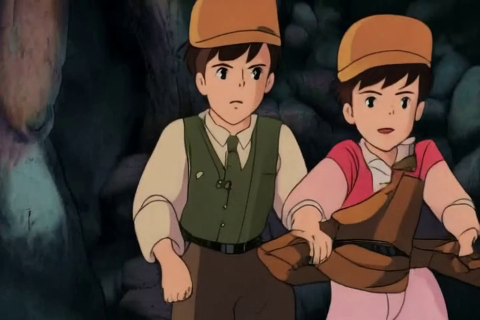} &
    \includegraphics[width=.11\textwidth]{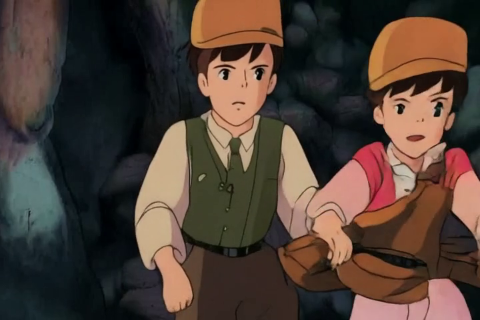} &
    \includegraphics[width=.11\textwidth]{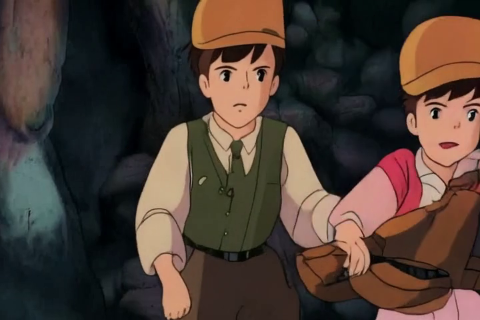} &
    \includegraphics[width=.11\textwidth]{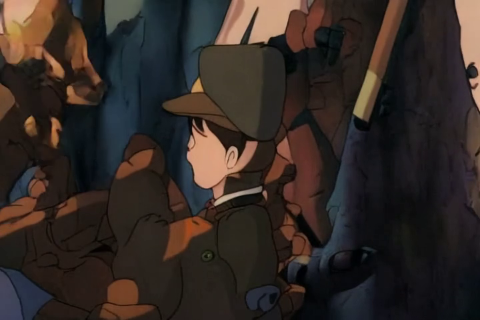} &
    \includegraphics[width=.11\textwidth]{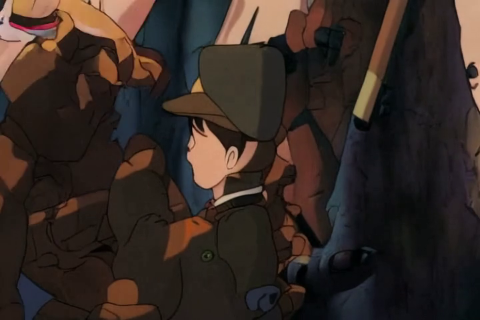} &
    \includegraphics[width=.11\textwidth]{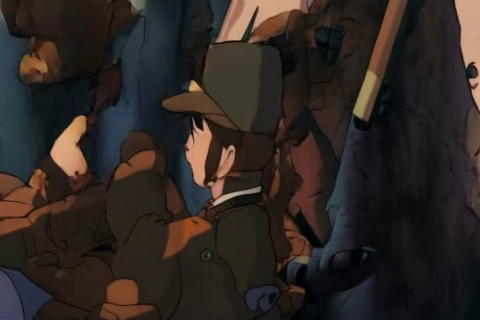} &
    \includegraphics[width=.11\textwidth]{images/main-figure-state/5.pdf}\\
  \multicolumn{4}{c}{\vspace{-14.8pt}} \\
    \raisebox{1.7\height}{\rotatebox[origin=c]{90}{\fontsize{7}{10}\selectfont GFC}} &
    \includegraphics[width=.11\textwidth]{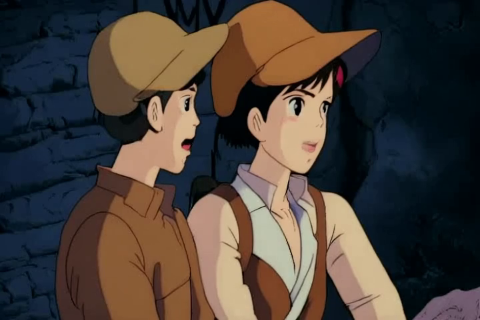} &
    \includegraphics[width=.11\textwidth]{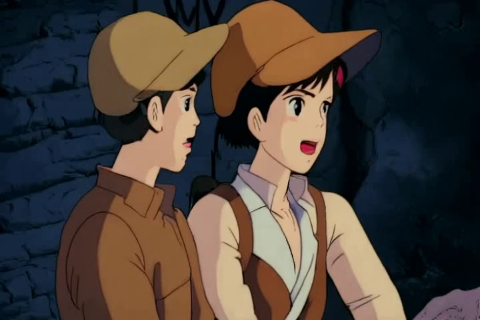} &
    \includegraphics[width=.11\textwidth]{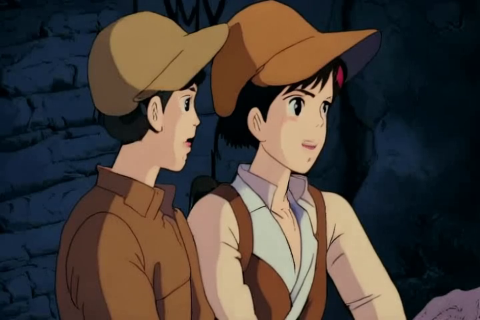} &
    \includegraphics[width=.11\textwidth]{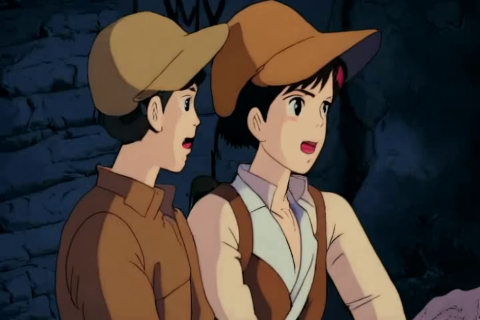} &
    \includegraphics[width=.11\textwidth]{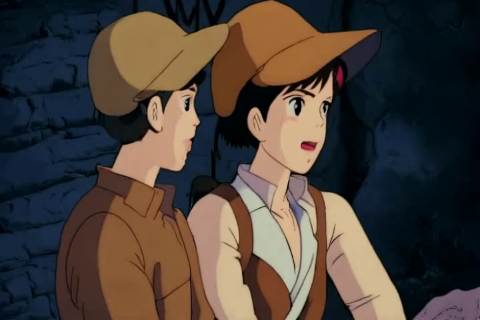} &
    \includegraphics[width=.11\textwidth]{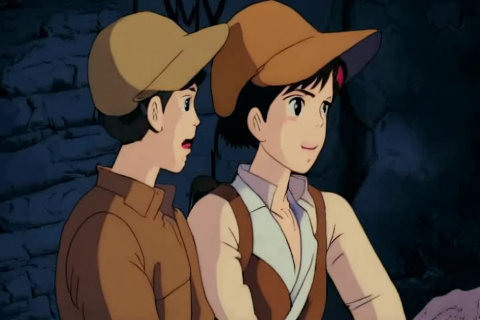} &
    \includegraphics[width=.11\textwidth]{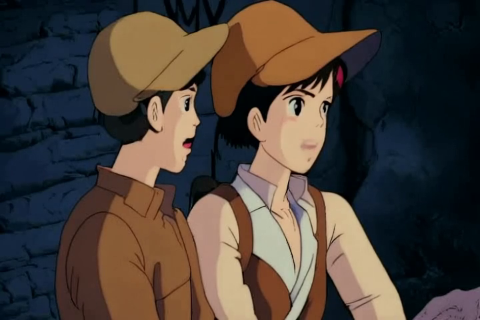} &
    \includegraphics[width=.11\textwidth]{images/main-figure-state/5.pdf} \\
  \end{tabular}
  \vspace{-3mm}
\caption{Visualization of infinite anime life simulations.}
  \label{fig: main visualization}
\end{figure*}

We compare infinite anime life simulation results\footnote{See Appendix F for character images and more qualitative results.} of AnimeGamer with GC and GFC in Figure~\ref{fig: main visualization}. GC and GFC neglect historical visual information, leading to a deficiency in contextual consistency. Additionally, they underperform at generalize interactions between characters from different anime films (the two characters in rounds 1 and 2 are from two distinct anime films) and character actions (in round 3, the action of ``flying on a broomstick" is exclusive to Qiqi in the training set). In contrast, AnimeGamer considers multimodal context in the generation process, thus delivering a more coherent and immersive game experience. Moreover, the generalization ability of the MLLM makes AnimeGamer perform well in character-centric commands. The tuning-free method GSC fails to achieve character consistency, which is crucial for the gaming experience, thus is unsuitable for this task.

\subsection{Ablation Study}

We randomly select one anime film within our dataset to conduct ablation studies. See Appendix C for details.

\begin{figure}[!t]
  \centering
  \setlength{\tabcolsep}{1pt} 
  \begin{tabular}{ccccc}
  \multicolumn{4}{c}{\makebox[0pt][c]{\small \hspace{7em} \raisebox{-0.3ex}{\includegraphics[height=1em]{images/icon/profile.png}} \hspace{0.2em} \textcolor[HTML]{FF4967}{Qiqi} \hspace{1em} \raisebox{-0.2ex}{\includegraphics[height=1em]{images/icon/joystick.png}} \hspace{0.2em} Excitedly laugh \hspace{1em} \raisebox{-0.3ex}{\includegraphics[height=1em]{images/icon/map.png}} \hspace{0.2em} Sky}} \\
     \raisebox{1.1\height}{\rotatebox[origin=c]{90}{\fontsize{7}{10}\selectfont w/o adapt}} &
    \includegraphics[width=.11\textwidth]{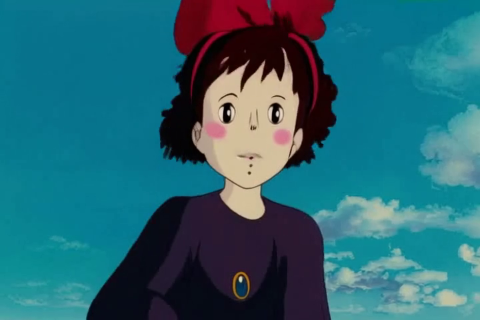} &
    \includegraphics[width=.11\textwidth]{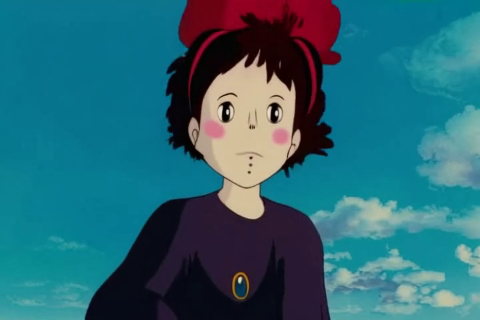} &
    \includegraphics[width=.11\textwidth]{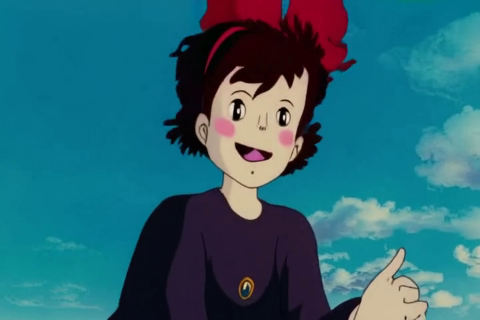} &
    \includegraphics[width=.11\textwidth]{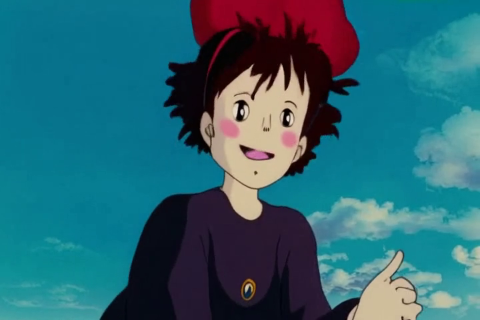} \\
    \multicolumn{4}{c}{\vspace{-14pt}} \\
    \raisebox{1.4\height}{\rotatebox[origin=c]{90}{\fontsize{7}{10}\selectfont w/ $\mathcal{L}_{\text{cos}}$}} &
    \includegraphics[width=.11\textwidth]{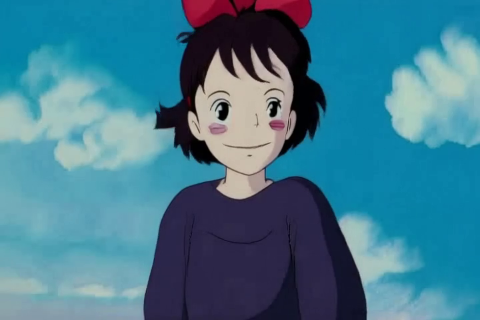} &
    \includegraphics[width=.11\textwidth]{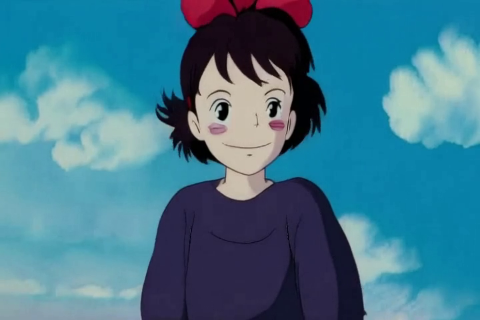} &
    \includegraphics[width=.11\textwidth]{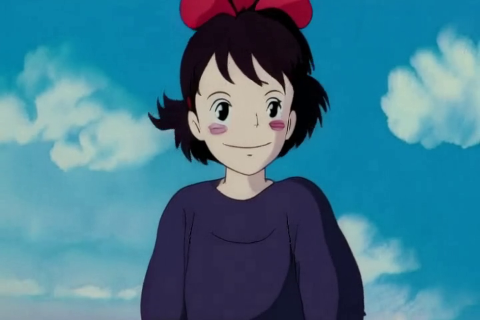} &
    \includegraphics[width=.11\textwidth]{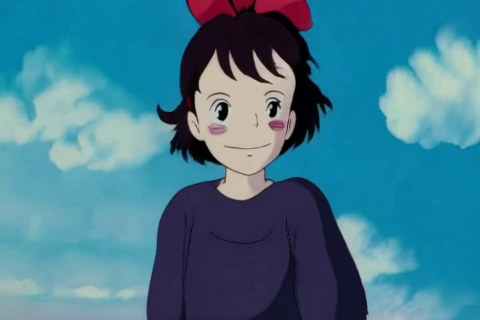} \\
    \multicolumn{4}{c}{\vspace{-14pt}} \\
    \raisebox{1.7\height}{\rotatebox[origin=c]{90}{\fontsize{7}{10}\selectfont \textbf{Ours}}} &
    \includegraphics[width=.11\textwidth]{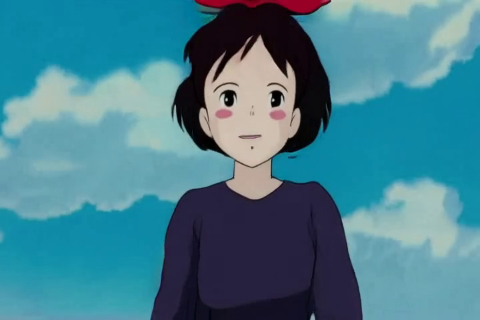} &
    \includegraphics[width=.11\textwidth]{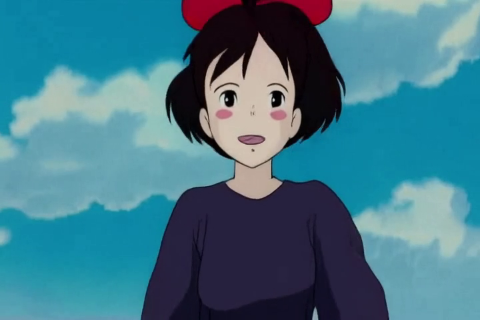} &
    \includegraphics[width=.11\textwidth]{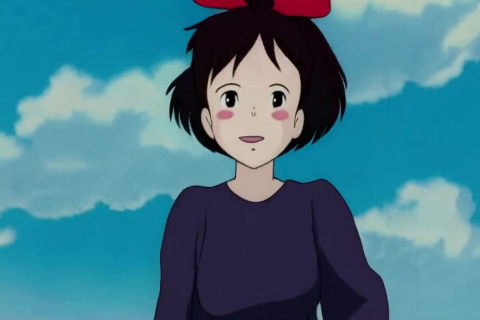} &
    \includegraphics[width=.11\textwidth]{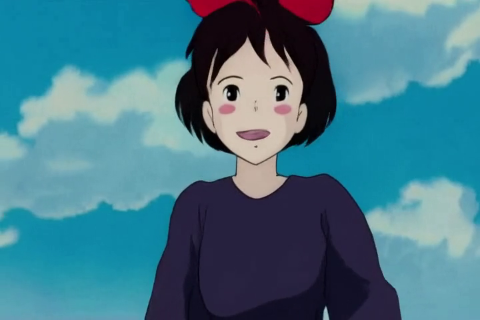} \\
  \end{tabular}
  \vspace{-3mm}
\caption{Visualization of ablation study on game state prediction and decoder adaptation. Our method outperforms in terms of character consistency and movement following.}
  \label{fig: ablation-2}
\end{figure}

\begin{figure}[!t]
  \centering
  \setlength{\tabcolsep}{1pt} 
  \begin{tabular}{ccccc}
  \multicolumn{4}{c}{\makebox[0pt][c]{\small \hspace{7em} \raisebox{-0.3ex}{\includegraphics[height=1em]{images/icon/profile.png}} \hspace{0.2em} \textcolor[HTML]{D0A19C}{Ursala} \hspace{1em} \raisebox{-0.2ex}{\includegraphics[height=1em]{images/icon/joystick.png}} \hspace{0.2em} Quickly read paper \hspace{1em} \raisebox{-0.3ex}{\includegraphics[height=1em]{images/icon/map.png}} \hspace{0.2em} Home}} \\
     \raisebox{0.78\height}{\rotatebox[origin=c]{90}{\fontsize{7}{10}\selectfont Ground Truth}} &
    \includegraphics[width=.11\textwidth]{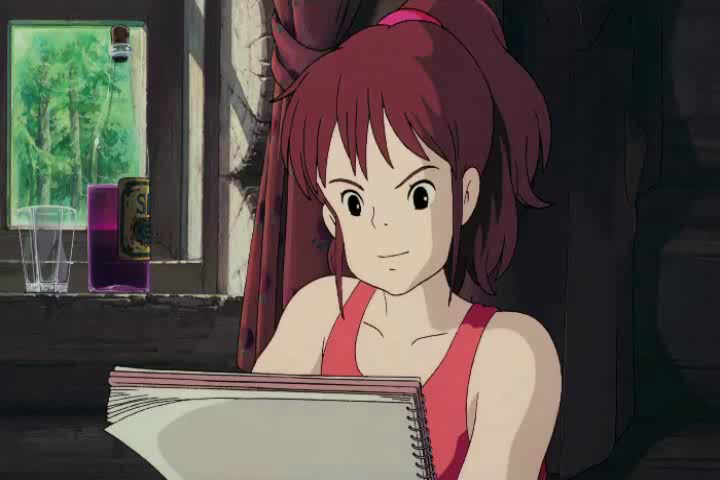} &
    \includegraphics[width=.11\textwidth]{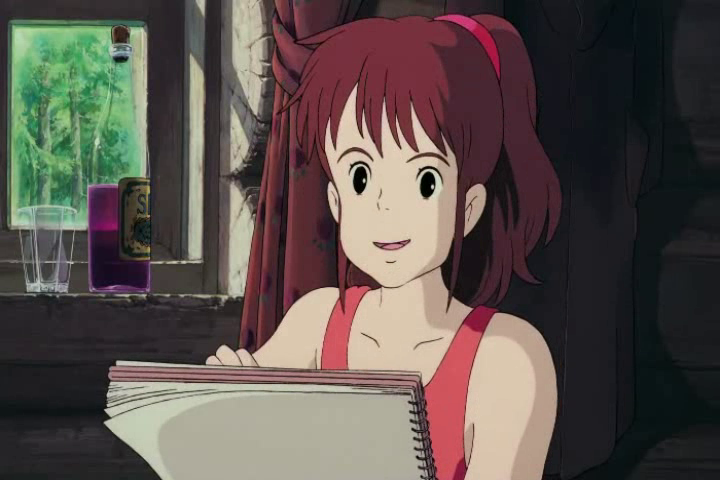} &
    \includegraphics[width=.11\textwidth]{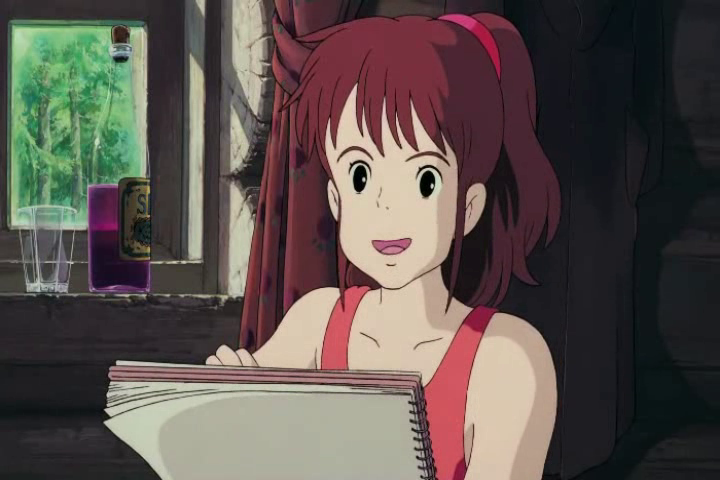} &
    \includegraphics[width=.11\textwidth]{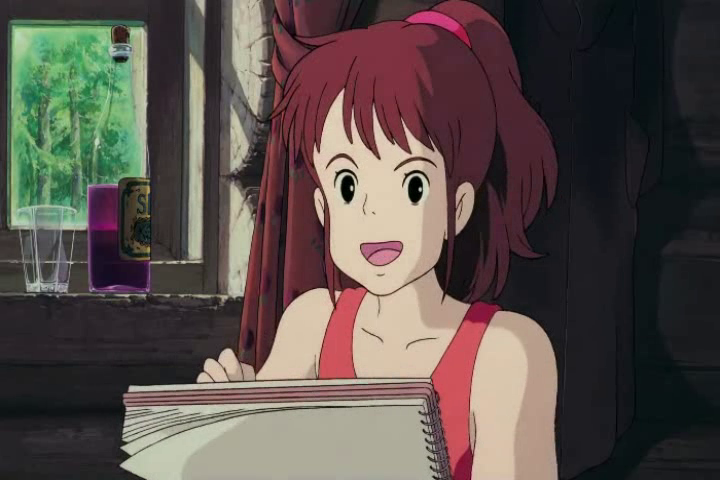} \\
    \multicolumn{4}{c}{\vspace{-12pt}} \\
    \raisebox{0.69\height}{\rotatebox[origin=c]{90}{\fontsize{7}{10}\selectfont w/ rand. frame}} &
    \includegraphics[width=.11\textwidth]{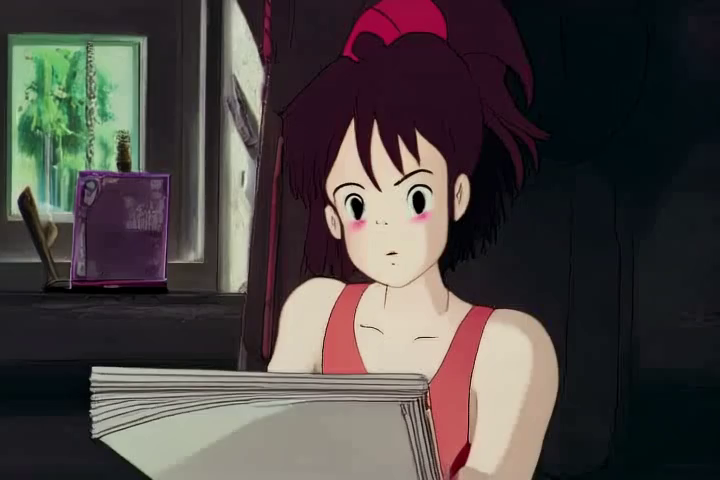} &
    \includegraphics[width=.11\textwidth]{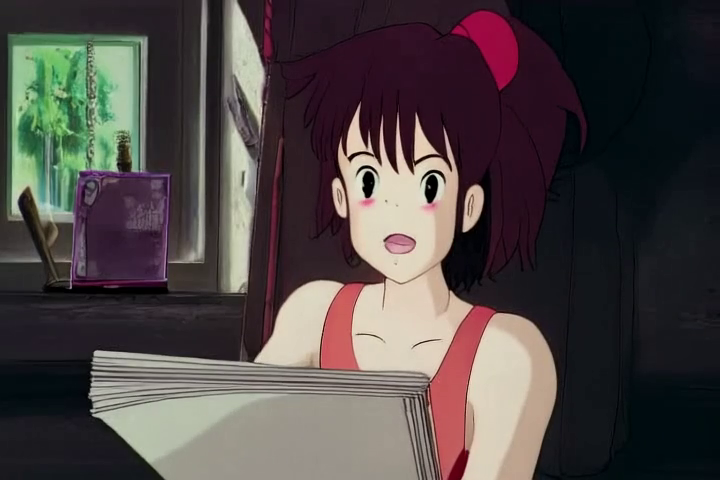} &
    \includegraphics[width=.11\textwidth]{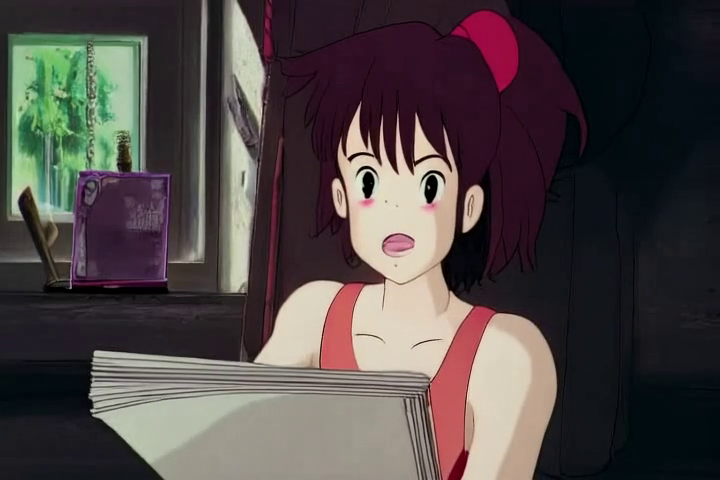} &
    \includegraphics[width=.11\textwidth]{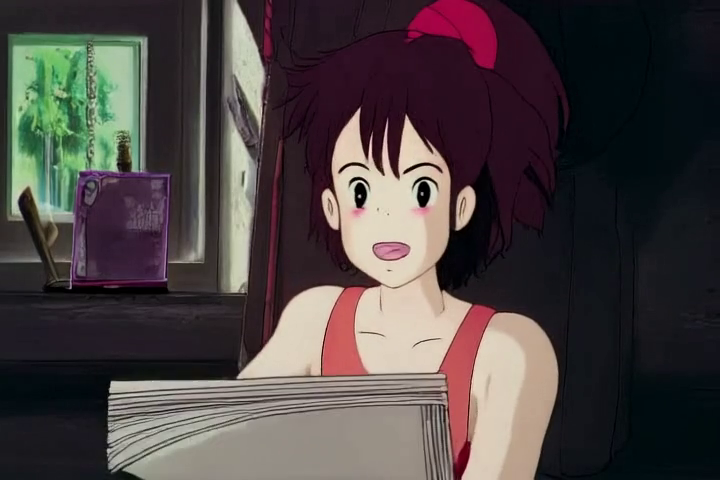} \\
    \multicolumn{4}{c}{\vspace{-12pt}} \\
    \raisebox{0.85\height}{\rotatebox[origin=c]{90}{\fontsize{7}{10}\selectfont w/ less para}} &
    \includegraphics[width=.11\textwidth]{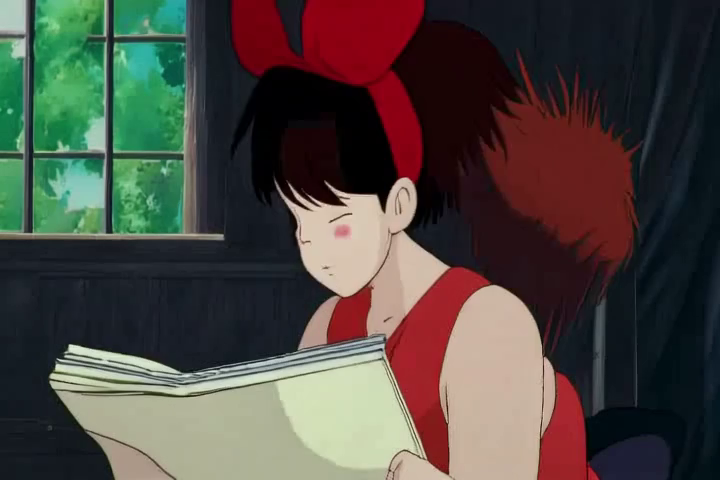} &
    \includegraphics[width=.11\textwidth]{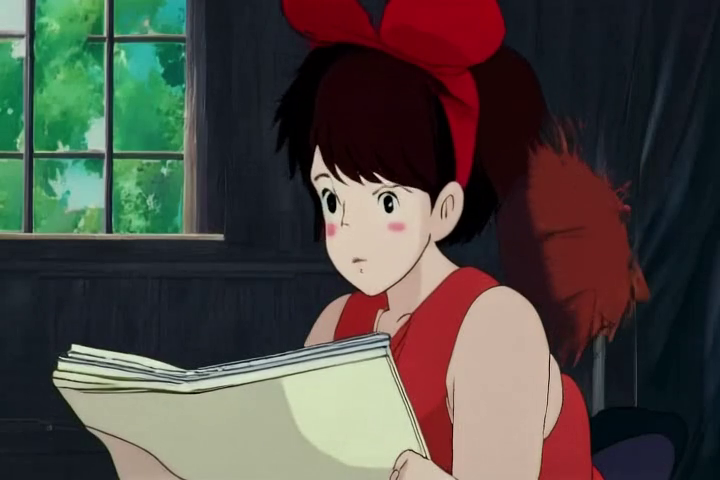} &
    \includegraphics[width=.11\textwidth]{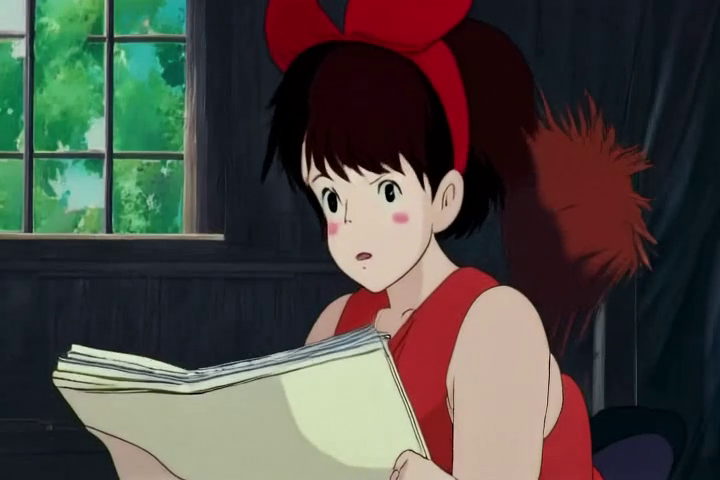} &
    \includegraphics[width=.11\textwidth]{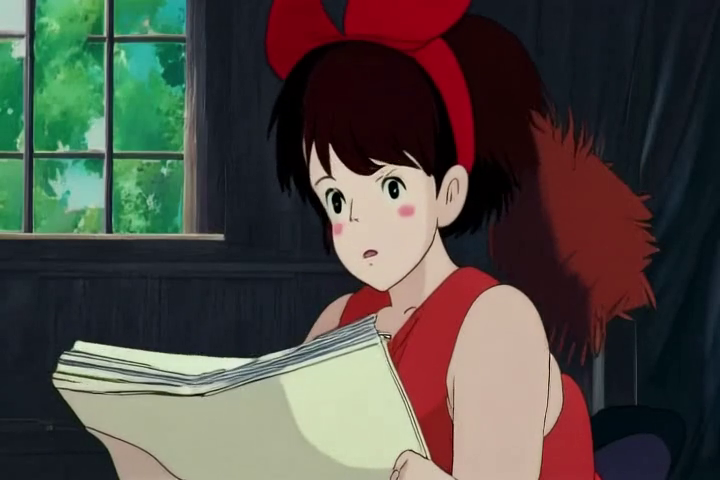} \\
    \multicolumn{4}{c}{\vspace{-12pt}} \\
    \raisebox{0.9\height}{\rotatebox[origin=c]{90}{\fontsize{7}{10}\selectfont w/ addition}} &
    \includegraphics[width=.11\textwidth]{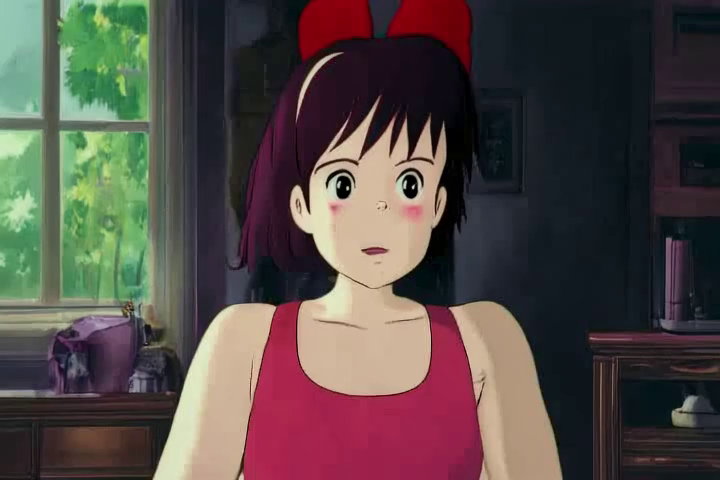} &
    \includegraphics[width=.11\textwidth]{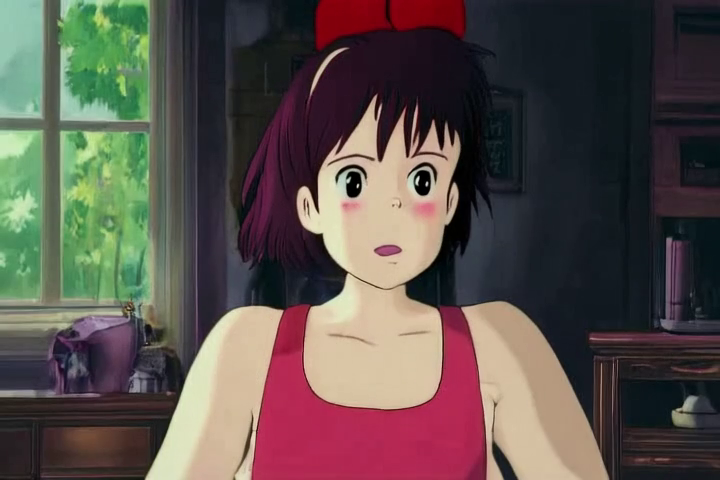} &
    \includegraphics[width=.11\textwidth]{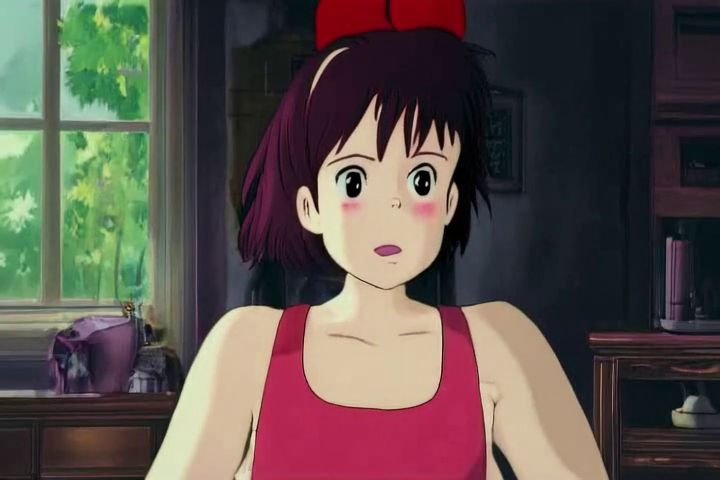} &
    \includegraphics[width=.11\textwidth]{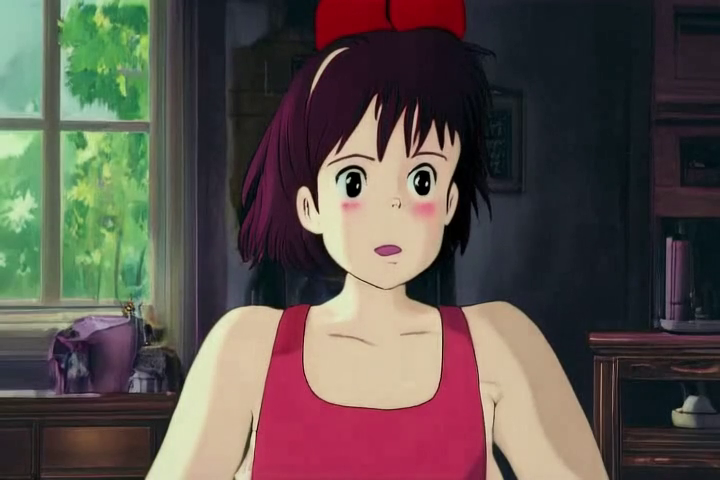} \\
    \multicolumn{4}{c}{\vspace{-12pt}} \\
    \raisebox{0.8\height}{\rotatebox[origin=c]{90}{\fontsize{7}{10}\selectfont w/ cross-attn}} &
    \includegraphics[width=.11\textwidth]{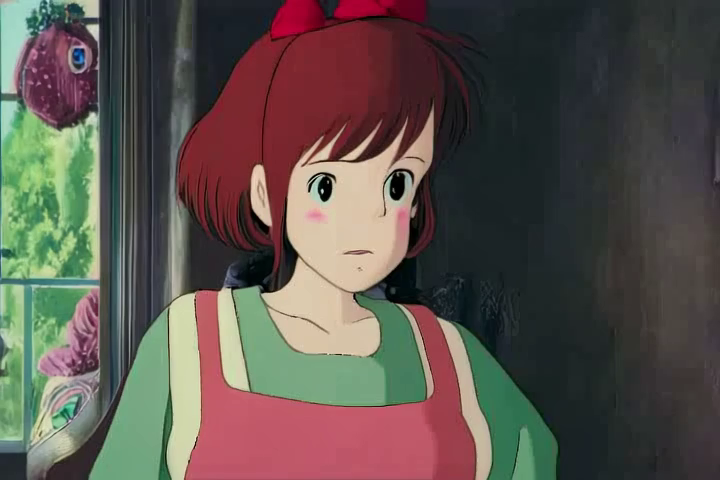} &
    \includegraphics[width=.11\textwidth]{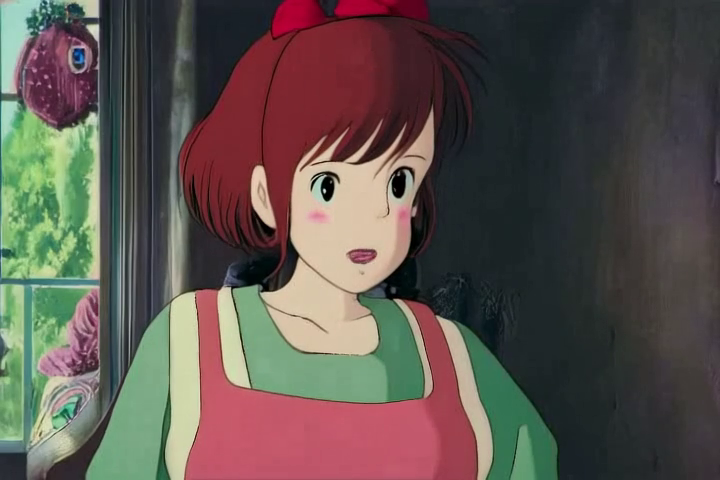} &
    \includegraphics[width=.11\textwidth]{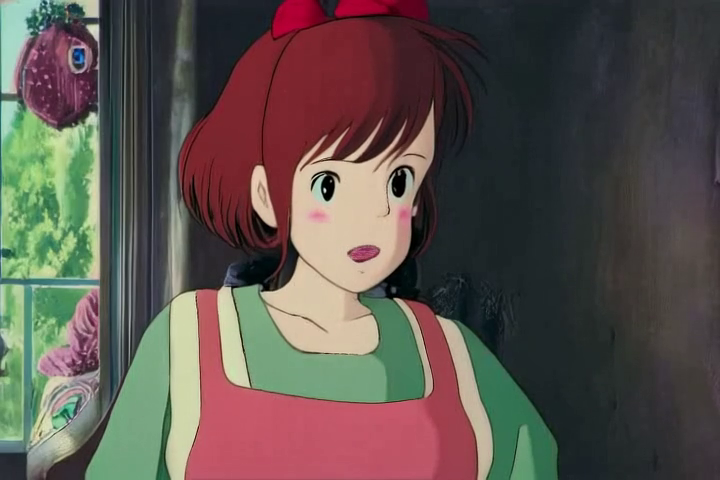} &
    \includegraphics[width=.11\textwidth]{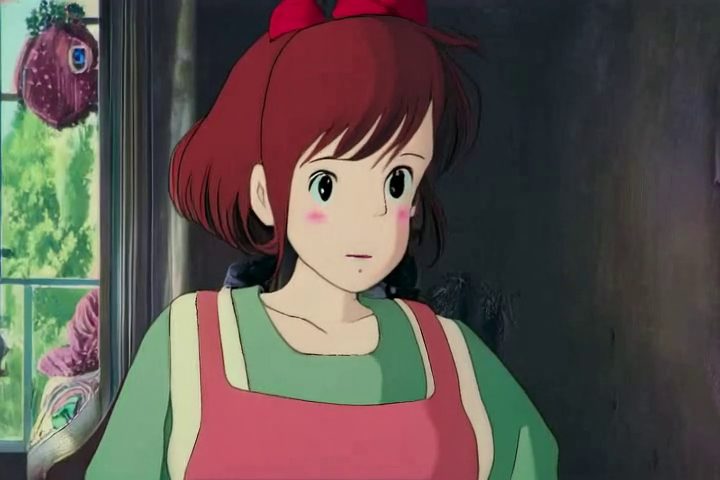} \\
    \multicolumn{4}{c}{\vspace{-12pt}} \\
    \raisebox{0.8\height}{\rotatebox[origin=c]{90}{\fontsize{7}{10}\selectfont w/o warm-up}} &
    \includegraphics[width=.11\textwidth]{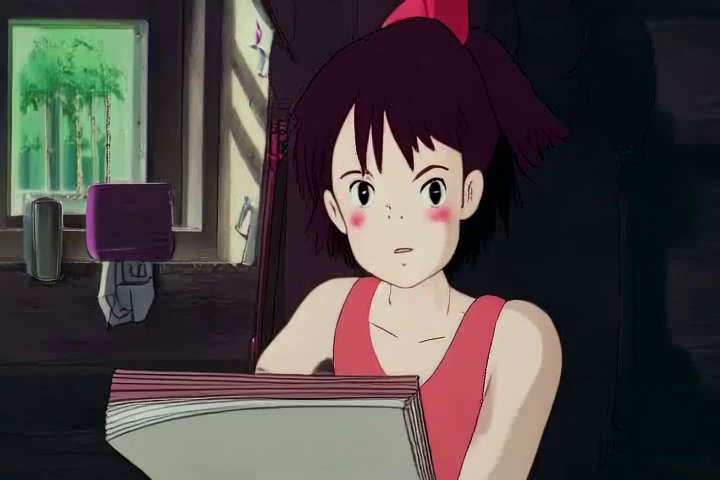} &
    \includegraphics[width=.11\textwidth]{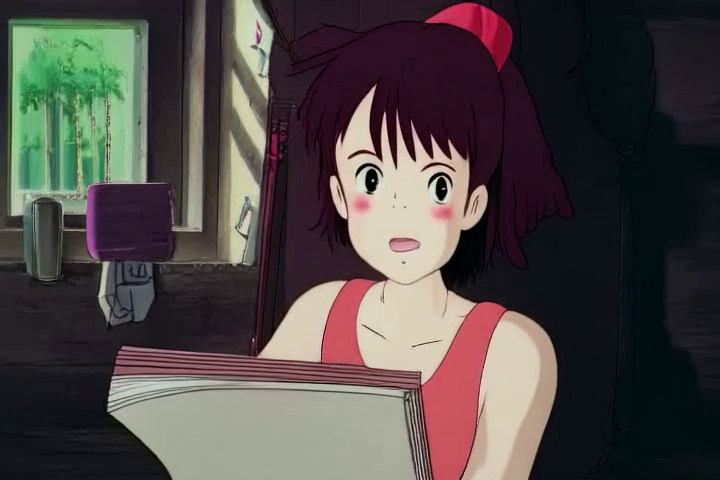} &
    \includegraphics[width=.11\textwidth]{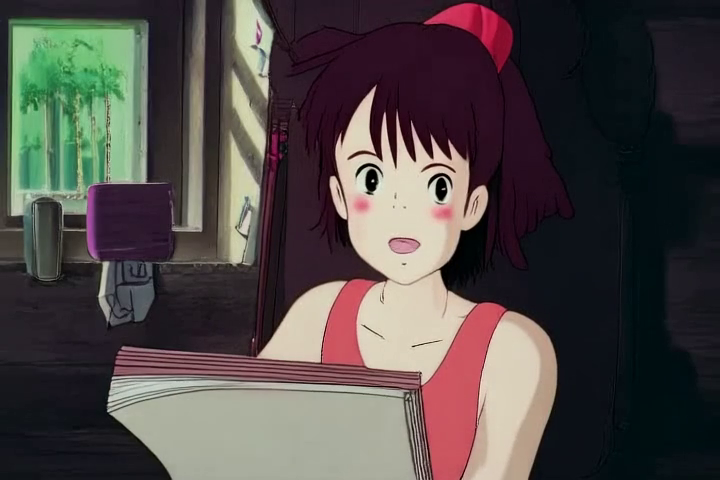} &
    \includegraphics[width=.11\textwidth]{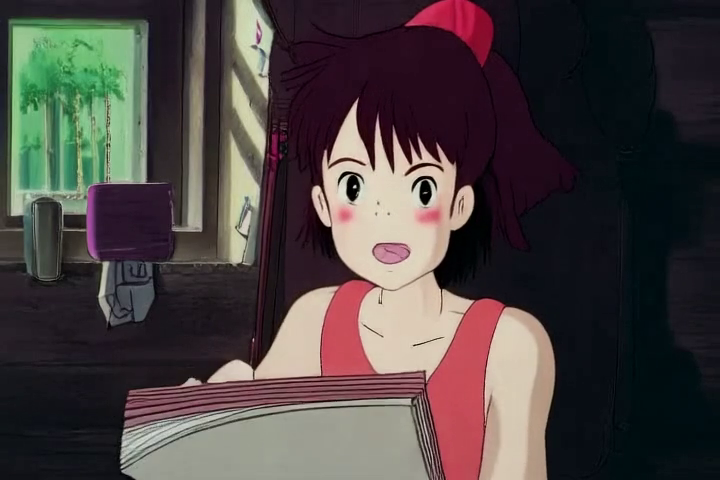} \\
    \multicolumn{4}{c}{\vspace{-12pt}} \\
    \raisebox{1.1\height}{\rotatebox[origin=c]{90}{\fontsize{7}{10}\selectfont w/o $f_{ms}$}} &
    \includegraphics[width=.11\textwidth]{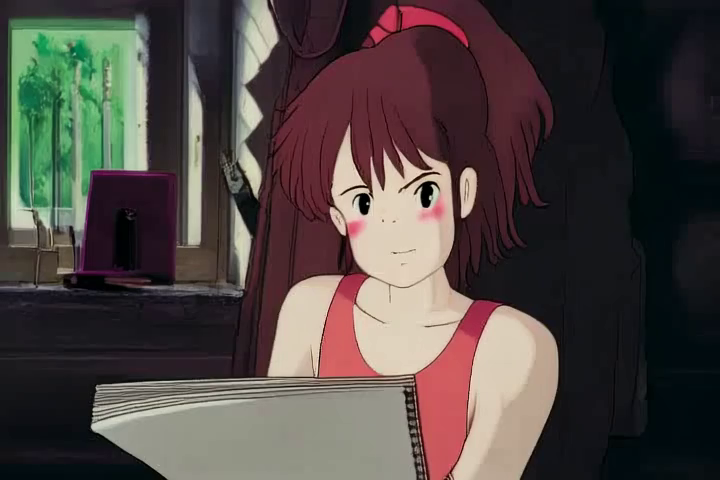} &
    \includegraphics[width=.11\textwidth]{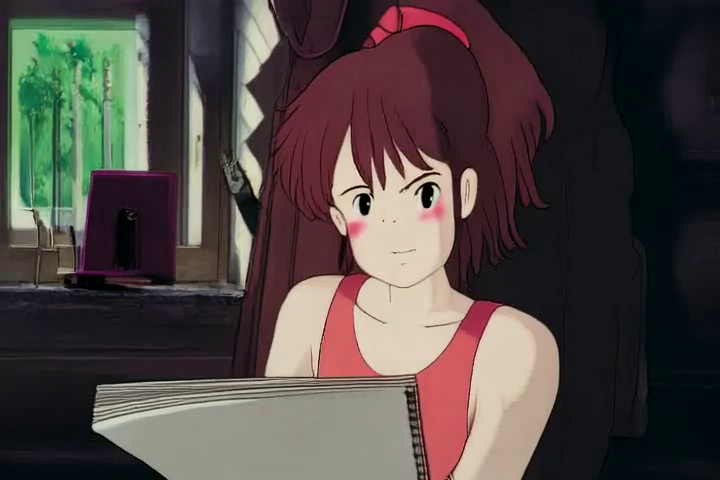} &
    \includegraphics[width=.11\textwidth]{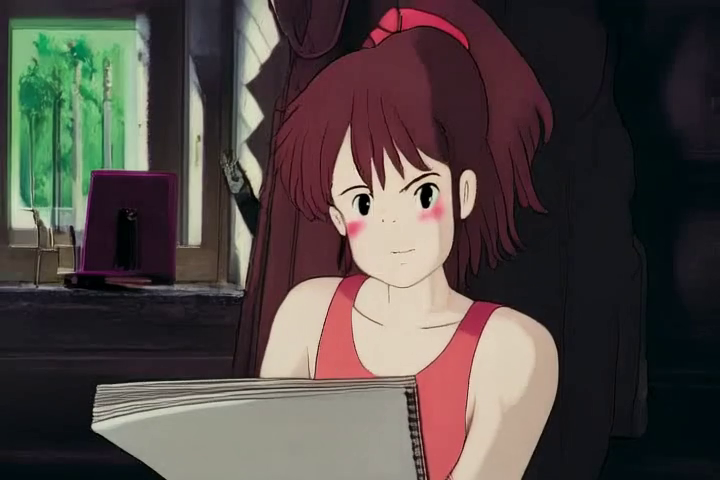} &
    \includegraphics[width=.11\textwidth]{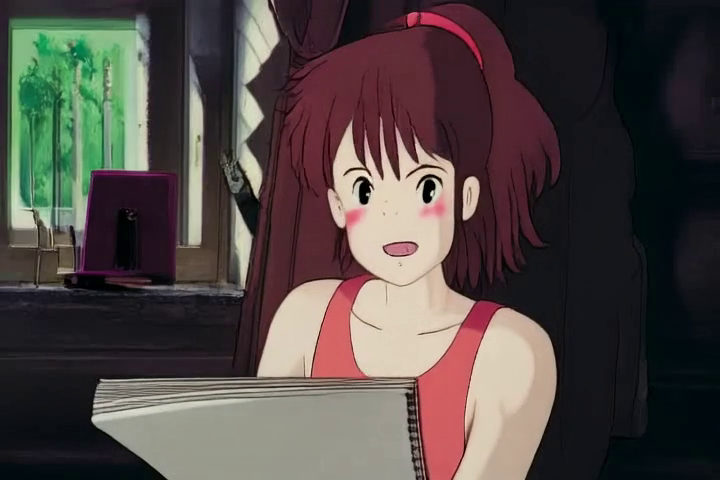} \\
    \multicolumn{4}{c}{\vspace{-12pt}} \\
    \raisebox{1.4\height}{\rotatebox[origin=c]{90}{\fontsize{7}{10}\selectfont \textbf{Ours}}} &
    \includegraphics[width=.11\textwidth]{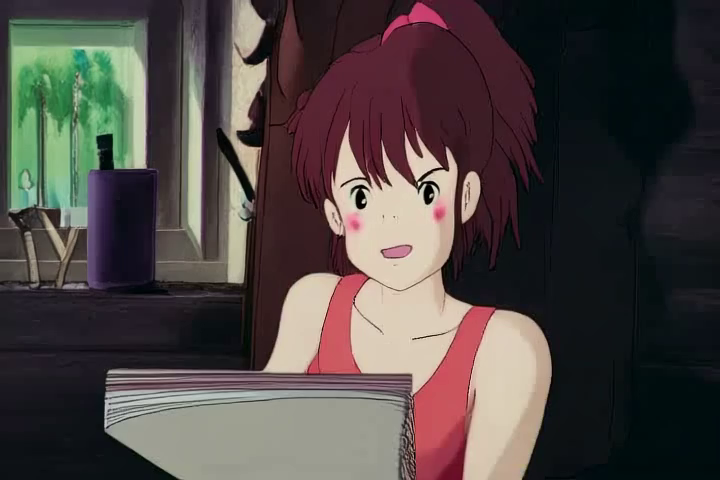} &
    \includegraphics[width=.11\textwidth]{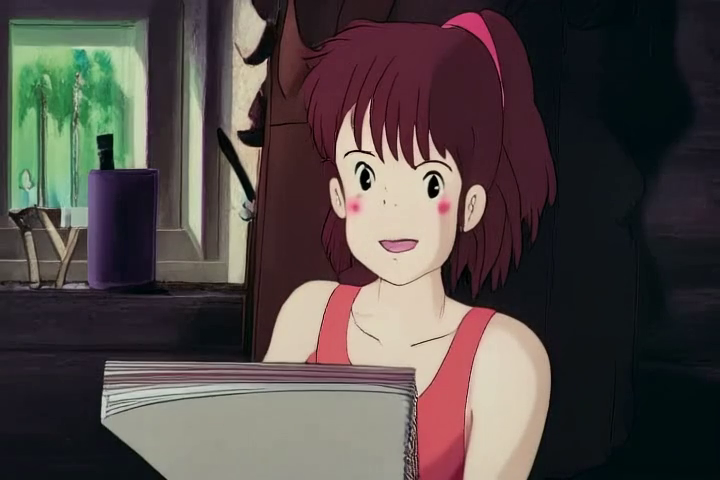} &
    \includegraphics[width=.11\textwidth]{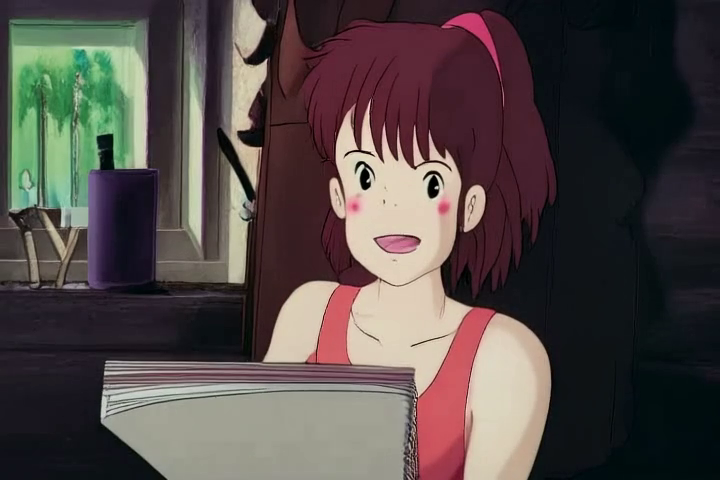} &
    \includegraphics[width=.11\textwidth]{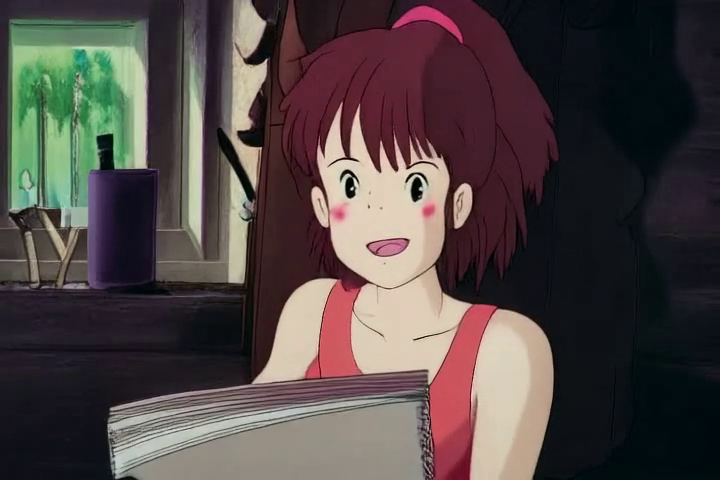} \\
  \end{tabular}
  \vspace{-3mm}
\caption{Results of ablation study on anime shot tokenization and de-tokenization. Our method outperforms in terms of image consistency and movement following.}
  \label{fig: ablation-s1}
\end{figure}

\begin{table}[!t]
\caption{Results of the ablation study of our AnimeGamer, where the columns in the table above pertain to the ablation experiments on the tokenizer and de-tokenizer (w/o MLLM).}
  \vspace{-5pt}
  \label{tab: Ablaiton}
  \centering
\resizebox{0.5\textwidth}{!}{
\begin{tabular}{lccccccc}
\toprule
\multicolumn{1}{c}{\multirow{2}{*}{Name}}  & \multicolumn{1}{c}{\multirow{2}{*}{w/o MLLM}} & \multicolumn{2}{c}{Image Consistency} & \multicolumn{2}{c}{Semantic Consistency} & 
\multicolumn{2}{c}{Motion Quality} \\ 
\cmidrule(lr){3-4}  \cmidrule(lr){5-6} \cmidrule(lr){7-8}

& \multicolumn{1}{c}{} & CLIP-I  & DreamSim  & CLIP-T & $\text{CLIP-T}^{\text{E}}$  & ACC-F  & MAE-F \\
\midrule
w/ rand. frame & \checkmark & 0.8446   & 0.4500   & 0.2480   & 0.2270  & 0.4744 & 0.5620 \\
w/ less para & \checkmark & 0.8406    & 0.4481    & 0.2450       & 0.2274     & 0.3649   & 0.6934  \\
w/ addition  & \checkmark   & 0.7684    & 0.6173    & 0.2311      & 0.232      & 0.4671   & 0.6058  \\
w/ cross-attn & \checkmark    & 0.7264    & 0.7084    & 0.2328      & 0.2333     & 0.3284   & 0.8102  \\
w/o warm-up & \checkmark    & 0.8306    & 0.5107    & 0.2447      & 0.2382     & 0.7028   & 0.4582  \\
w/o $f_{ms}$ & \checkmark & 0.8533    & 0.6894    & 0.2829      & 0.2518     & 0.1824   & 1.2189  \\
\cellcolor[HTML]{E6F0E8}\textbf{Ours} & \cellcolor[HTML]{E6F0E8}\checkmark &\cellcolor[HTML]{E6F0E8}\textbf{0.8672}   &\cellcolor[HTML]{E6F0E8}\textbf{0.7928}   &\cellcolor[HTML]{E6F0E8}\textbf{0.2831}     &\cellcolor[HTML]{E6F0E8}\textbf{0.2523}    &\cellcolor[HTML]{E6F0E8}\textbf{0.7293}  &\cellcolor[HTML]{E6F0E8}\textbf{0.4029} \\
\midrule
w/o adapt   & \ding{55}   & 0.6831    & 0.4937    & 0.1889      & 0.1898     & 0.3649   & 0.8467  \\
w/ $\mathcal{L}_{\text{cos}}$   & \ding{55}  & 0.7628    & 0.5966    & \textbf{0.2228}      & 0.2117     & 0.6649   & 0.4467  \\
\cellcolor[HTML]{E6F0E8}\textbf{Ours}  & \cellcolor[HTML]{E6F0E8}\ding{55} &\cellcolor[HTML]{E6F0E8}\textbf{0.7856}   &\cellcolor[HTML]{E6F0E8}\textbf{0.6084}   &\cellcolor[HTML]{E6F0E8}0.2212     &\cellcolor[HTML]{E6F0E8}\textbf{0.2203}    &\cellcolor[HTML]{E6F0E8}\textbf{0.6722}  &\cellcolor[HTML]{E6F0E8}\textbf{0.4883} \\
\bottomrule
\end{tabular}
}
\end{table}

\vspace{1mm} \noindent \textbf{Ablation on Animation Shot Tokenization and De-tokenization.} $\mathcal{E}_a$ plays a crucial role in encoding anime clips into action-aware multimodal representations. We demonstrate its efficiency by comparing it with four variants\footnote{Detailed in Appendix C.}, as well as removing the warm-up training phase. The results presented in Table~\ref{tab: Ablaiton} and Figure~\ref{fig: ablation-s1} indicate that reducing the learnable parameters of $\mathcal{E}_a$ or combining $f_v$ with $f_{md}$ via element-wise addition or cross-attention leads to a decline in all metrics and visual quality. This can be attributed to the disruption of spatial positional information within the visual feature. Additionally, using random frame instead of first frame to obtain $f_v$ or remove the warm-up training phase also leads to a decrease in consistency between the generated animation shot and the reference character. This can be caused by the increased difficulty in training. Finally, we remove $f_{ms}$ in $\mathcal{D}_a$, which results in a decline in motion control quality. This indicates that relying solely on text to control the motion scope is unreliable.

\vspace{1mm} \noindent \textbf{Ablation on Next Game State Prediction.} Some studies~\cite{xiao2025videoauteur} employ Cosine Similarity Loss for MLLM when training to fit continuous features. We adopt this approach in our AnimeGamer, denoted as ``w/ $\mathcal{L}_{\text{cos}}$". Results in Table~\ref{tab: Ablaiton} and Figure~\ref{fig: ablation-2} show that the impact of ${\mathcal{L}_\text{cos}}$ is marginal.

\vspace{1mm} \noindent \textbf{Ablation on Decoder Adaptation.} We conduct ablation by removing the decoder adaptation training phase, denoted as ``w/o adapt". As illustrated in Table~\ref{tab: Ablaiton} and Figure~\ref{fig: ablation-2}, removing the decoder adaptation training phase leads to artifacts in the generated videos. These disadvantages may lead to unsatisfactory gaming experiences.

\vspace{-1mm}
\section{Conclusion and Limitation}

In this paper, we propose AnimeGamer for infinite anime life simulation. Users can continuously interact with the game world as anime characters through open-ended language instructions. AnimeGamer generates multi-turn game states that consist of dynamic animation shots and updates to character states, including stamina, social, and entertainment values. Through modeling animation shots using action-aware multimodal representations, we train a MLLM to predict the next animation shot representations by taking the history instructions and multimodal representations as the input.  Evaluation through both automated metrics and human evaluation shows that AnimeGamer outperforms baseline methods across various gaming aspects. 

Our focus has been on developing an effective method for transforming characters into interactive, playable entities within infinite games, without further exploration of the extension to open domains. Our task setting aligns with the most recent work in infinite game generation, which emphasizes training models with custom characters and evaluating them in closed domains. In future work, we will explore the generalization to unseen characters.

\clearpage
\appendix
\section*{Overview}
In this appendix, we present the following:
\begin{itemize}   
    \item Details of our AnimeGamer in Section \ref{appdix: detail of game mllm}.
    \item Dataset Construction Pipeline in Section \ref{appdix: Dataset Construction Details}.
    \item Implementation details of AnimeGamer and other baselines in Section \ref{appdix: Implementation details}.
    \item Details of the evaluation benchmark in Section \ref{appdix: Evaluation Benchmark Construction}.
    \item Human Evaluation in Section \ref{appdix: Human Evaluation}.
    \item Additional visualization results in Section \ref{appdix: Additional results}.
\end{itemize}

\section{Details of AnimeGamer}
\label{appdix: detail of game mllm}

\vspace{1mm} \noindent \textbf{Special Tokens in MLLM.} We add special token to the tokenizer of our MLLM to generate game states and formulate the output. Specifically, we use \texttt{\textless MS\textgreater \textless /MS\textgreater} to represent the start and end of motion scope, \texttt{\textless ST\textgreater \textless /ST\textgreater} for stinema value, \texttt{\textless SC\textgreater \textless /SC\textgreater} for social value, \texttt{\textless ET\textgreater \textless /ET\textgreater} for entertainment value. To continuous generate the action-aware multimodal representations, We add 226 learnable query tokens \texttt{\textless IMAGE i\textgreater} to stimulate continuous generation, and \texttt{\textless VS\textgreater \textless /VS\textgreater} to represent start and end of the animation shot representation.

\vspace{1mm} \noindent \textbf{Motion Scope.} We employ Memflow~\cite{dong2024memflow} to compute the optical flow transformation for each frame within the video. Subsequently, we convert them into absolute values to represent the motion scope. A filtering threshold of $r$=0.2 is adopted to filter out the background information. After that, we calculate the average value of the remaining part to denote the motion scope of an animation shot. Next, we divide the range into five levels, which serve as discrete targets for the MLLM to fit.

\section{Dataset Construction Details}
\label{appdix: Dataset Construction Details}

\vspace{1mm} \noindent {\bf Video Pre-processing.} Taking an anime film as an example, we first download the film and crop its borders. Then, we resize it to the corresponding size and divide it into several segments using a scene - detection model~\cite{chen2024panda}. Next, each segment is split according to a fixed time period (2 seconds). In this way, we obtain the video training data arranged in timestamp order. In addition, we download the reference images of each protagonist to locate the characters in each video segment.

\begin{figure*}[!t]
	\centering
	\includegraphics[width=1\textwidth]{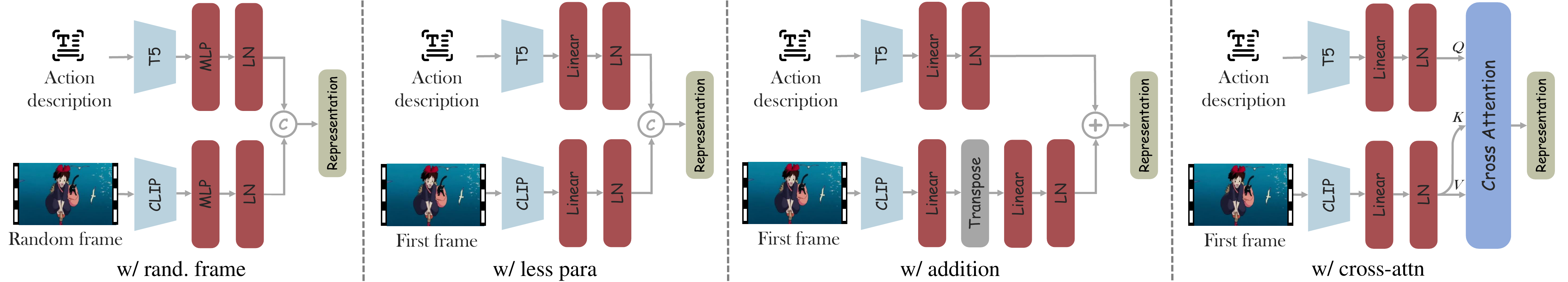}
	\caption{Four variants of our animation shot encoder. 1) We use random frame instead of first frame to obatain the action-aware multimodal representation, denoted as ``w/ rand. frame"; 2) We replace the MLP module in $\mathcal{E}_a$ with a single Linear layer to reduce learnable parameters, denoted as ``w/ less para"; 3) We combine $f_v$ with $f_{md}$ via element-wise addition, denoted as ``w/ addition"; 4) We combine $f_v$ with $f_{md}$ via cross-attention, denoted as ``w/ cross-attn".}
	\label{fig: model-ablation}
\end{figure*}

\vspace{1mm} \noindent {\bf Captioning.} We utilize Intern-VL-26B~\cite{chen2024internvl} to generate captions for each animation shot. We input the protagonist reference images and eight evenly-sampled frames from an anime clip as visual input. To match the character, we employ the following prompt:
\begin{tcolorbox}
Image-1: \textless image\textgreater  Image-2: \textless image\textgreater  
According to the characters' index in Image-1, your task is to answer how many characters are in Image-2 (4 frames from a anime video) and what their indices are. Response format: '[Num]\textless number of characters\textgreater [ID]\textless index of the character\textgreater '; Response example: '[Number]2[Index]1,3'. If no characters are detected, please respond with '[Number]0[Index]0'. Your response:
\end{tcolorbox}
To acquire descriptions of motion, the environment, and character states, we utilize the following prompt:
\begin{tcolorbox}
Image-1: \textless image\textgreater  Image-1 is from an anime clip, Your task is to extract a structured description based on the information.  First, I will give you the movement level \textless ML\textgreater  respectively. The movement level is categorized into five levels: Level 1: very small movement amplitude, almost imperceptible; Level 2: small movement amplitude, slight swaying or adjustments; Level 3: moderate movement amplitude, appropriate movement or adjustments; Level 4: large movement amplitude, noticeable and significant; Level 5: very large movement amplitude, extremely obvious and intense. Next, you need to generate the following information: (1) subject \textless S\textgreater , motion description \textless MD\textgreater  and environment \textless EV\textgreater. Please use a single word for subject and background, use a simple phrase for motion in the present simple tense. (2) Movement adverb \textless MA\textgreater : Based on \textless ML\textgreater , give \textless MD\textgreater  a fitting adverb. (3) Social interaction \textless SC\textgreater : If there are two or more characters interacting socially in the scene, such as talking, hugging, walk together or kissing, use 1; otherwise, use 0 to indicate no social action. (4) Entertainment \textless ET\textgreater : If the protagonist is engaged in entertainment activities, sports or relaxing, such as reading, riding, flying, swimming, whispering, archery... use 1; otherwise, use 0 to indicate no entertainment action.
(5) Stamina \textless ST\textgreater : Stamina can be restored through actions like eating, drinking, lying, sleeping, hugging, treatment... If the \textless MD\textgreater  are restoring stamina, fill in 1; otherwise, use -1. Example output: \textless S\textgreater Girl\textless /S\textgreater \textless MD\textgreater run\textless /MD\textgreater \textless EV\textgreater Forest

\textless /EV\textgreater \textless MA\textgreater slowly\textless /MA\textgreater \textless SC\textgreater 0\textless /SC\textgreater \textless ET\textgreater 

1\textless /ET\textgreater \textless ST\textgreater -1\textless /ST\textgreater. Your response:
\end{tcolorbox}

\section{Implementation Details}
\label{appdix: Implementation details}

In this section, we present the implementation details of our AnimeGamer in Section~\ref{app: 1}, baseline methods in Section~\ref{app: 2} and the ablation studies in Section~\ref{app: 3}. 

\subsection{AnimeGamer} 
\label{app: 1}

\vspace{1mm} \noindent \textbf{Animation Shot Encoding and Decoding.} In this phase, we initialize the parameters of our animation shot decoder using CogvideoX-2B\footnote{THUDM/CogVideoX-2b}. We apply LoRA to the 3D-Attention with a rank of 64. The learning rate is set to 2e-4. To enhance generalization capabilities, we initially pretrain using 100k samples from the WebVid~\cite{bain2021frozen}. Subsequently, we start with a warm-up phase where $\mathcal{E}_a$ s trained for 10,000 steps. This is followed by a joint training phase of $\mathcal{E}_a$ and $\mathcal{D}_a$ which extends for an additional 80,000 steps.

\vspace{1mm} \noindent \textbf{Next Game State Prediction.} For the MLLM, we initialize our model with the weight of Mistral-7B and train it using LoRA, facilitated by the peft library. The LoRA rank is set to 32, with lora-alpha also set to 32. The learning rate is 5e-5, and the training is carried out for 15,000 steps.

\vspace{1mm} \noindent \textbf{Decoder Adaptation.} In this stage, we fine-tune only $\mathcal{D}_a$. The learning rate is 5e-5, and the training is executed for 10,000 steps.

\subsection{Baselines}
\label{app: 2}
\vspace{1mm} \noindent \textbf{GSC.} We utilize StoryDiffusion~\cite{zhou2024storydiffusion} based on SDXL~\cite{podell2023sdxl}, where the instructions for a 10-round game are input simultaneously to generate the corresponding images. Then, we use the Cogvideox-5B-I2V\footnote{THUDM/CogVideoX-5b-I2V} model to convert these images into animation shots. During this process, action instructions are provided as prompts to the pretrained I2V model.

\vspace{1mm} \noindent \textbf{GFC.} We fine-tune the T2I model FlUX\footnote{black-forest-labs/FLUX.1-dev} using LoRA. For training, we pair the first frame of each animation shot with its corresponding instruction to form image-text pairs. We employ LoRA with a rank of 32 and train for 200,000 steps. During testing, we convert images to video using the same method as in GSC.

\vspace{1mm} \noindent \textbf{GC.} We fine-tune the CogvideoX-2B model using LoRA, employing the same configuration as used for training $\mathcal{D}_a$.

\subsection{Ablation Study}
\label{app: 3}

In the ablation study, we randomly selected a movie ``Qiqi's Delivery Service" from the training dataset as the training data. We split approximately 2,000 training samples into a training set and a test set with an 8:2 ratio. The ablation on animation shot tokenization and de-tokenization does not incorporate the MLLM, in order to focus on the reconstruction ability of $\mathcal{E}_a$ and $\mathcal{D}_a$ for animation shots.

\vspace{1mm} \noindent \textbf{Ablation on Animation Shot Tokenization and De-tokenization.} We construct four variants for $\mathcal{E}_a$, as shown in Figure~\ref{fig: model-ablation}. 1) We use a random frame instead of the first frame as $f_v$, denoted as ``w/ rand. frame". 2) We replace the MLP with a simpler Linear layer to align features, denoted as ``w/ less para". 3) We use element-wise addition to unify $f_v$ and $f_{md}$, denoted as ``w/ addition". 4) We use cross-attention to unify $f_v$ and $f_{md}$, denoted as ``w/ cross-attn".

\vspace{1mm} \noindent \textbf{Ablation on Next Game State Prediction.} In this ablation study, we incorporate the Cosine Loss into the training process. The overall training loss is a combination given by:
\begin{equation}
\mathcal{L}=\mathcal{L}_{\text{CE}}+\alpha\mathcal{L}_{\text{MSE}}+\beta\mathcal{L}_{\text{cos}},
\label{eq: loss}
\end{equation}
where the hyperparameters $\alpha$ and $\beta$ are set to 0.5.

\section{Evaluation Benchmark Construction}
\label{appdix: Evaluation Benchmark Construction}

\vspace{1mm} \noindent \textbf{MLLM as benchmark constructor.} We use the following prompt for GPT-4o to generate our evaluation benchmark.

\begin{tcolorbox}
You are a world model for an anime life simulation. You can generate stories of the character living in the world. The stories should sound like a game and leave space for user interaction. Now, you need to generate a 10-panel story (simulation game) with [Character] as the main character. For each turn, you should generate the following components: 1) Characters \textless S\textgreater : The main character must appear in each panel, and 0-1 additional characters can be included as supporting characters, chosen from [Characters]. 2) Motion Description \textless MD\textgreater : Describe the main character's action with a simple phrase. 3) Environment \textless EV\textgreater : Describe the current environment with one word. 4) Main character's state: you need to generate the following information: (1) Motion Level \textless ML\textgreater : The movement level is categorized into number 1-5: 1: very small movement amplitude 2: small movement amplitude, slight swaying or adjustments 3: moderate movement amplitude, appropriate movement or adjustments 4: large movement amplitude, noticeable and significant 5: very large movement amplitude, extremely obvious and intense (2) Movement adverb \textless MA\textgreater : Based on \textless ML\textgreater , give \textless MD\textgreater  a fitting adverb. (3) Social interaction \textless SC\textgreater : If there are two or more characters interacting socially in the scene, such as talking, hugging, walking together, or kissing, use 1; otherwise, use 0 to indicate no social action. (4) Entertainment \textless ET\textgreater : If the protagonist is engaged in entertainment activities, sports, or relaxing, such as reading, riding, flying, swimming, whispering, archery, use 1; otherwise, use 0 to indicate no entertainment action. (5) Stamina \textless ST\textgreater : Stamina can be restored through actions like eating, drinking, lying, sleeping, hugging, treatment... If the \textless MD\textgreater  are restoring stamina, use 1; otherwise, use -1. For the entire story, here are some instructions you need to follow: 1) Ensure continuity between different panels as much as possible. Encourage different actions in the same scene or return to a previous scene in subsequent panels. 2) Keep it realistic and as close to a life simulation game scenario as possible. Please use common scenes and easily representable actions, and avoid including tiny, difficult-to-generate objects. 3) Output format: Each line represents one turn, using the following format: \textless S\textgreater Characters\textless /S\textgreater \textless MD\textgreater Motion 
Description\textless /MD\textgreater \textless EV\textgreater Environment\textless /EV\textgreater \textless ML

\textgreater Motion Level\textless /ML\textgreater \textless MA\textgreater Movement adverb

\textless /MA\textgreater \textless SC\textgreater Social interaction\textless /SC\textgreater \textless ET\textgreater 

Entertainment\textless /ET\textgreater \textless ST\textgreater Stamina\textless /ST\textgreater 
\end{tcolorbox}

\vspace{1mm} \noindent \textbf{MLLM as a judge.} We use the following prompt for GPT-4o to assess the output of the models.

\begin{tcolorbox}
Please act as an impartial judge and evaluate the quality of the generation story video contents provided by N AI agents. Here's some instructions you need to follow:

1) Story Composition: Each story consists of 5 scenes, and I will provide you with their respective prompts.

2) Evaluation: For each AI agent's output, I will present you with an image composed of 5 frames extracted from the videos. The image in the i-th row represent 5 frames extracted from the generated video corresponding to scene i.

3) Evaluation Criteria: You need to score each AI agent's output based on Overall Quality \textless OA\textgreater: The overall gaming experience. Text Alignment \textless TA\textgreater: The alignment between the prompt and the generated results. Contextual Coherence \textless ConC\textgreater: Whether the content of each scene can connect naturally, Character Consistency \textless ChaC\textgreater: Are the characters in each scene consistent with the provided reference characters? Emotional Consistency \textless EC\textgreater: The consistency between the expression of the scenes and the emotional statements in the prompt. Visual Coherence \textless VC\textgreater: Are the colors, styles, and compositions of the scenes consistent? The score range for these criteria is from 1 to 10, with higher scores indicating better overall performance. 4) Output Format: Your output should contain four lines, each starting with the evaluation criteria code such as \textless OA\textgreater, followed by N numbers representing the scores for each of the N agents, separated by spaces. Finally, provide a brief explanation of your evaluation on a new line. 5) Evaluation Requirements: Avoid any bias, ensure that the order of presentation does not affect your decision. Do not let the length of the response influence your evaluation. Do not favor certain agent names. 
\end{tcolorbox}

\section{Human Evaluation}
\label{appdix: Human Evaluation}
For the human evaluation, we recruit 20 participants who hold at least a bachelor's degree and have prior experience in image or video generation. A total of 9-round games with 50 samples are presented to the participants. We showcase the animation shots and character states generated by various models to the participants in the form of a PowerPoint presentation and ask them to fill out an Excel spreadsheet. They are required to rate the performance of different models for each metric in every game. Subsequently, we convert the rankings into absolute scores: 10 points for the first-ranked model, 7 points for the second, 4 points for the third, and 1 point for the fourth. Finally, we calculate the average performance of each model.

\section{Additional Qualitative results}
\label{appdix: Additional results}
\begin{figure}[h]
  \centering
  \setlength{\tabcolsep}{3pt} 
  \begin{tabular}{ccc}
  Ponyo & Sosuke & Qiqi \\
    \includegraphics[width=0.11\textwidth,height=0.11\textwidth]{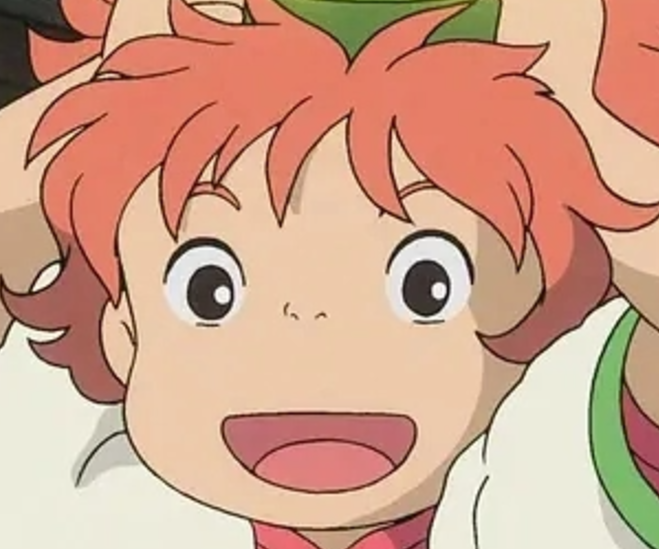} &
    \includegraphics[width=0.11\textwidth,height=0.11\textwidth]{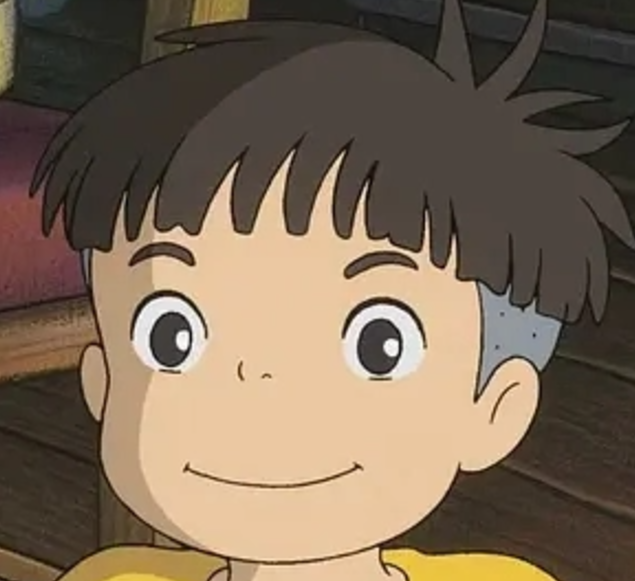} &
    \includegraphics[width=0.11\textwidth,height=0.11\textwidth]{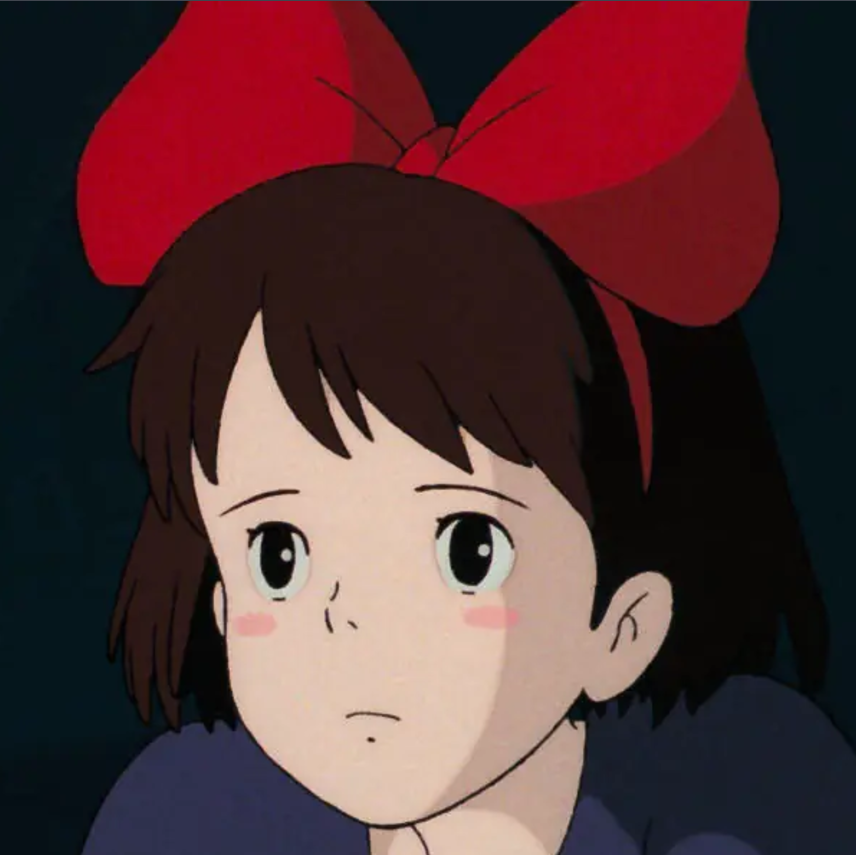} \\
    \multicolumn{3}{c}{\vspace{-12pt}} \\
  Firefly & Pazu & Sheeta \\
    \includegraphics[width=0.11\textwidth,height=0.11\textwidth]{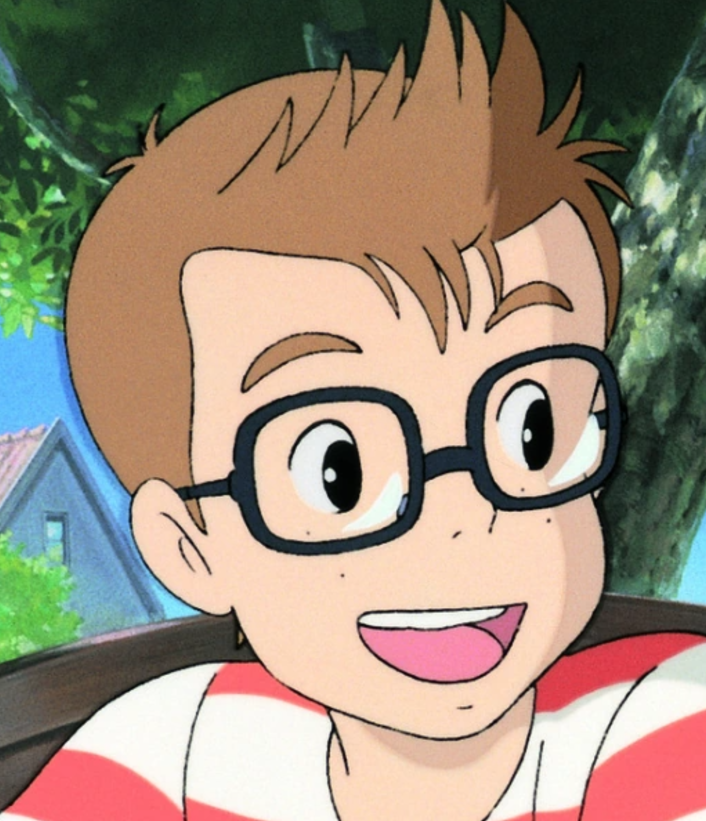} &
    \includegraphics[width=0.11\textwidth,height=0.11\textwidth]{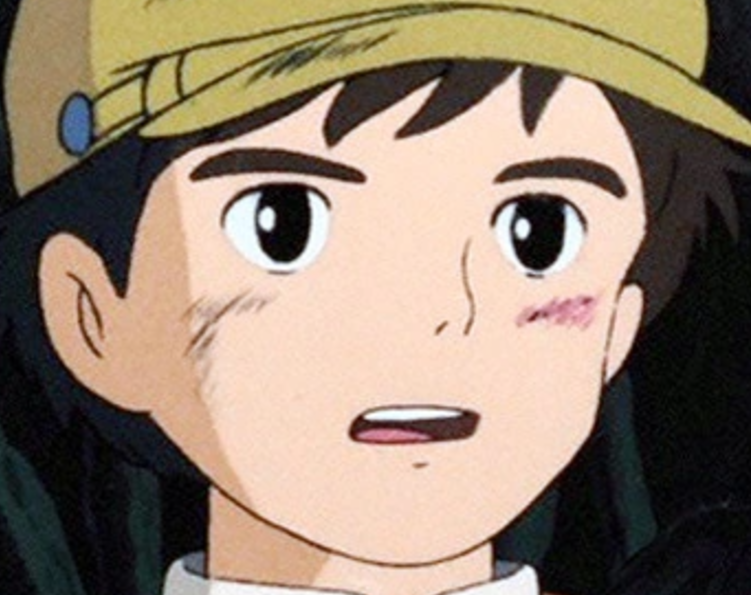} &
    \includegraphics[width=0.11\textwidth,height=0.11\textwidth]{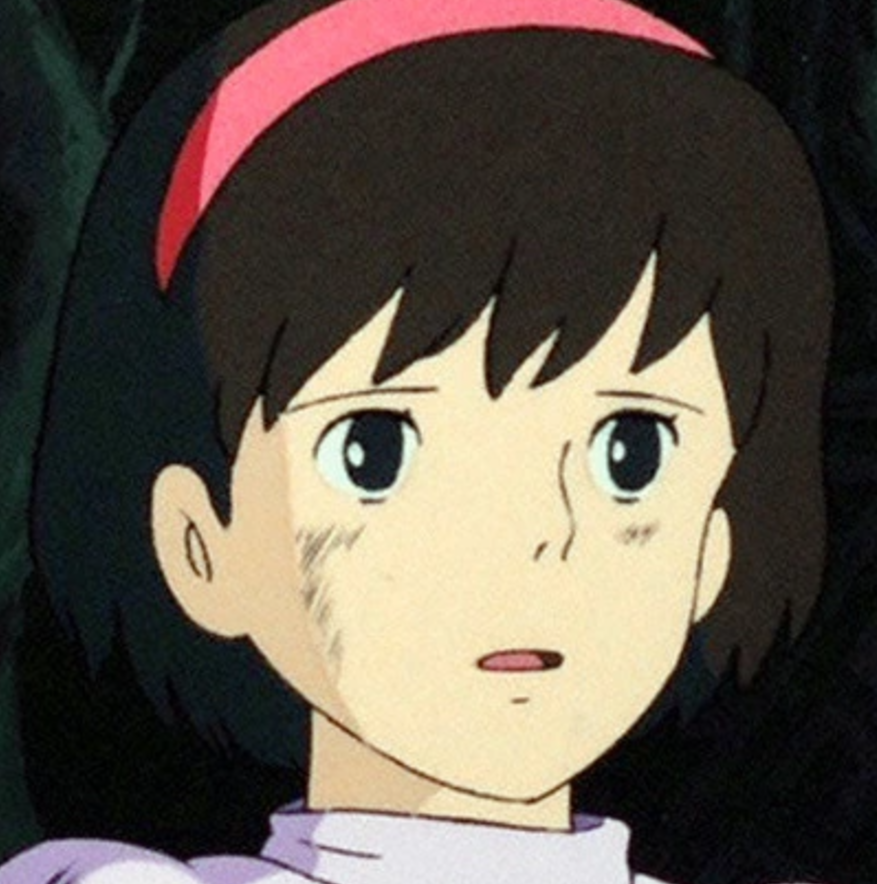}  \\
  \end{tabular}
  \vspace{-3mm}
\caption{Image of characters in the paper.}
  \label{fig: face}
\end{figure}

We present the image of characters appeared in our paper in Figure~\ref{fig: face}. We present the infinite game generation results of AnimeGamer and other baselines in our homepage: \url{https://howe125.github.io/AnimeGamer.github.io/}.

\clearpage
{
    \small
    \bibliographystyle{ieeenat_fullname}
    \bibliography{main}
}

\end{document}